 \documentclass[preprint, review,3p,times, 10pt,authoryear]{elsarticle}

\usepackage{amssymb}
\usepackage{pgf-pie}
\usepackage{longtable} 
\usepackage{graphicx}
\usepackage[hyphens]{url}
\usepackage{hyperref}
\hypersetup{colorlinks=true,breaklinks=true}
\usepackage{rotating} 
\usepackage{lscape} 
\usepackage{multirow} 
\usepackage{float} 
\usepackage{textcmds}  
\usepackage{romannum}
\usepackage{makecell}

\usepackage{lineno}

\journal{Artificial Intelligence in Agriculture, vol 6, pp. 138-155, 2022, \url{https://doi.org/10.1016/j.aiia.2022.09.002}}

\begin{document}

\begin{frontmatter}


\title{A Systematic Review of Machine Learning Techniques for Cattle Identification: Datasets, Methods and Future Directions}




\def\correspondingauthor{\footnote{Corresponding author: email@gmail.com}}
\author[1,4]{Md Ekramul Hossain}\ead{mdhossain@csu.edu.au}
\author[1,5,4]{Muhammad Ashad Kabir\corref{cora}}
\cortext[cora]{Corresponding author: School of Computing, Mathematics and Engineering, Charles Sturt University, Panorama Ave, Bathurst, NSW 2795. Ph.+61263386259, Email: akabir@csu.edu.au}
\ead{akabir@csu.edu.au}
\author[1,4]{Lihong Zheng}\ead{lzheng@csu.edu.au}
\author[5,2,4]{Dave L. Swain}\ead{dave.swain@terracipher.com}
\author[5,3,4]{Shawn McGrath}\ead{shmcgrath@csu.edu.au}
\author[5,4]{Jonathan Medway}\ead{jmedway@csu.edu.au}
\address[1]{School of Computing, Mathematics and Engineering, Charles Sturt University, Bathurst, NSW 2795, Australia}
\address[5]{Gulbali Institute for Agriculture, Water and Environment, Charles Sturt University, Wagga Wagga, NSW, 2678, Australia}
\address[2]{TerraCipher Pty. Ltd., Alton Downs, QLD 4702, Australia}
\address[3]{Fred Morley Centre, School of Animal and Veterinary Sciences, Charles Sturt University, Wagga Wagga, NSW 2678, Australia}
\address[4]{Food Agility CRC Ltd, Sydney, NSW 2000, Australia}


\begin{abstract}
Increased biosecurity and food safety requirements may increase demand for efficient traceability and identification systems of livestock in the supply chain. The advanced technologies of machine learning and computer vision have been applied in precision livestock management, including critical disease detection, vaccination, production management, tracking, and health monitoring. This paper offers a systematic literature review (SLR) of vision-based cattle identification. More specifically, this SLR is to identify and analyse the research related to cattle identification using Machine Learning (ML) and Deep Learning (DL). This study retrieved 731 studies from four online scholarly databases. Fifty-five articles were subsequently selected and investigated in depth. For the two main applications of cattle detection and cattle identification, all the ML based papers only solve cattle identification problems. However, both detection and identification problems were studied in the DL based papers. Based on our survey report, the most used ML models for cattle identification were support vector machine (SVM), k-nearest neighbour (KNN), and artificial neural network (ANN). Convolutional neural network (CNN), residual network (ResNet), Inception, You Only Look Once (YOLO), and Faster R-CNN were popular DL models in the selected papers. Among these papers, the most distinguishing features were the muzzle prints and coat patterns of cattle. Local binary pattern (LBP), speeded up robust features (SURF), scale-invariant feature transform (SIFT), and Inception or CNN were identified as the most used feature extraction methods. 
This paper details important factors to consider when choosing a technique or method. We also identified major challenges in cattle identification. There are few publicly available datasets, and the quality of those datasets are affected by the wild environment and movement while collecting data. The processing time is a critical factor for a real-time cattle identification system. Finally, a recommendation is given that more publicly available benchmark datasets will improve research progress in the future.

\end{abstract}




\end{frontmatter}


\section{Introduction} {\label{introduction}}
The demand for efficient traceability and identification systems for livestock is growing due to biosecurity and food safety requirements in the supply chain. The advanced technologies of machine learning and computer vision have been applied in precision livestock management, including critical disease detection, vaccination, production management, tracking, health monitoring, and animal well-being monitoring \citep{LR1awad2016classical}. {\lq{Cattle identification}\rq} refers to {\lq{cattle detection}\rq} and {\lq{cattle recognition}\rq} \citep{LR6mahmud2021systematic}. Cattle identification systems start from manual identification to automatic identification with the help of image processing. Traditional cattle identification systems such as ear tagging \citep{LR1awad2016classical}, ear notching \citep{INneary2002methods}, and electronic devices \citep{CIOruiz2011role} have been used for individual identification in cattle farming. Disadvantages of these individual identification methods include the possibility of losses, duplication, electronic device malfunctions, and fraud of the tag number \citep{INrossing1999animal,CIOroberts2006radio}. These are the issues and challenges for cattle identification in livestock farm management.

With the advent of computer-vision technology, cattle visual features have gained popularity for cattle identification \citep{AID1kusakunniran2018automatic,AID2andrew2016automatic,AID24andrew2017visual,AID35de2020recognition}. Visual feature based cattle identification systems are used to detect and classify different breeds or individuals based on a set of unique features. In recent years, machine learning (ML) and deep learning (DL) approaches have been widely used for automatic cattle identification using visual features \citep{AID2andrew2016automatic,AID16tharwat2014cattle,AID23andrew2019aerial,AID34qiao2019individual,li-plos-one2021}. ML and DL are subfields of artificial intelligence that can solve complex problems for automatic decision-making. ML is mainly divided into two approaches, such as supervised learning and unsupervised learning. The supervised ML approach is defined by its use of labelled datasets, whereas the unsupervised learning uses ML algorithms to analyse and cluster unlabeled datasets. An unsupervised ML approach can detect hidden patterns in data without human supervision \citep{INjaniesch2021machine}. DL approaches are useful in areas with large and high-dimensional datasets. Thus, DL models are usually outperformed over traditional ML models in the area of text, speech, image, video, and audio data processing \citep{INlecun2015deep}. There are two main steps in the development of ML and DL models. In the first step, a training dataset is used to train the model, and in the second, the model is validated using a separate validation dataset. Thus, a trained model is created that is later used on the test dataset to determine its performance based on the test dataset. The dataset used for ML models includes the features and their corresponding outcomes or labels. The features are extracted from the input data using a feature extraction method. DL algorithms can automatically extract high-level features from the dataset and learn from these features. Although the implementation of the ML and DL models is straightforward, there are some challenges with selecting algorithms, tuning parameters, and features for better prediction accuracy \citep{INjaniesch2021machine}.        

Several important review studies have been completed in livestock farm management. Some recent literature reviews have addressed various research challenges in livestock farming, such as identification, tracking, and health monitoring, using tag-based, ML, and DL approaches. Recently, \citet{LR1awad2016classical} and \citet{LR2kumar2020cattle} reviewed the literature on using different classical and visual biometrics methods for cattle identification and tracking. \citet{LR3li2021practices} reviewed the deep learning-based approaches for classification, object detection and segmentation, pose estimation, and tracking for different kinds of animals such as cattle, pigs, sheep, and poultry. A systematic literature review based on applying ML and DL approaches in precision livestock farming by \citet{LR4garcia2020systematic} focused on grazing and animal health. \citet{LR5qiao2021intelligent} summarised the ML and DL approaches in precision cattle farming for cattle identification, body condition score evaluation, and live weight estimation. They reviewed a small number of articles (n=13) related to cattle identification using ML and DL approaches. \citet{LR6mahmud2021systematic} conducted a systematic literature review showing the recent progress of DL applications for cattle identification and health monitoring. Their review included only a few articles related to cattle identification. Moreover, these review articles focused on the combination of different types of challenges (e.g., tracking, pose estimation, weight estimation, identification, and detection) solved by tag-based, ML, and DL methods in precision livestock farming. Thus, they lack in providing a comprehensive review on cattle identification. Also, the existing review articles lack information on ML and DL applications combined for cattle identification as they cover partly either ML or DL for cattle identification. Moreover, the details of the cattle dataset for identification are not discussed. In this context, an extensive systematic literature review is needed, particularly for the challenge of cattle identification addressed by ML and DL approaches. Also, the details of the dataset used in the relevant articles need to be discussed, and the current trend of using ML and DL techniques in cattle identification and future research opportunities with challenges need to be identified.

This systematic literature review (SLR) aims to summarise and analyse the ML and DL applications used extensively in cattle identification. A total of 55 articles for cattle identification and detection have been selected for this SLR. The reviewed articles are first summarised, and then the datasets used in the selected articles are discussed. We then analyse the reviewed articles for trends in using ML and DL approaches for cattle identification in recent years before presenting the feature extraction methods and performance evaluation metrics extracted from the reviewed articles. Finally, the challenges and future research directions in this field are discussed.   

\section{Methodology} {\label{methods}}

\subsection{Review process}
The review process of an SLR is divided into three phases -- planning, conducting, and reporting the review \citep{MTDkitchenham2007guidelines}. In the first phase, the research questions for the SLR are identified. Based on the research questions, the electronic databases and search terms or keywords were determined. The search keywords are used to create a search string that is applied to the different electronic databases to extract the related articles for the SLR. This study used the IEEE Xplore, Science Direct, Scopus, and Web of Science databases. These databases were selected to cover a wide range of studies in our targeted sector as they index most of the journals from various publishers such as Springer, ACM, Inderscience, Elsevier, Sage, Taylor \& Francis, IOS, Wiley, and so on. In the second phase, the relevant research studies are identified by searching the databases. After that, the selection criteria are determined for the quality assessment of the primary studies. The eligible studies are selected by applying the selection criteria, and then the relevant data are extracted from the selected articles based on the research questions. In the final phase, the extracted data are analysed and used to address the research questions. Then, the results are reported in the form of tables and figures followed by a brief discussion of research challenges and future research opportunities.

\subsection{Research questions}\label{sec:rq}
This SLR focuses on published research studies into cattle identification using ML and DL approaches. The search process identifies potential primary studies that address the research questions. The answers to the research questions are discussed based on the data extracted from the selected studies. This study defined the following seven research questions (RQs) for the SLR.    
 
\begin{itemize}
    \item RQ1: What ML models are used in cattle identification?
    \item RQ2: What DL models are used in cattle identification?
    \item RQ3: What datasets are used in cattle identification?
    \item RQ4: What feature extraction methods are used in cattle identification?
    \item RQ5: What performance evaluation metrics are used for ML and DL models in cattle identification?
    \item RQ6: What are the best  ML and DL models used in a specific cattle identification problem? 
    \item RQ7: What are the challenges in solving cattle identification problems?
\end{itemize}

\subsection{Search strategy}
A search strategy is applied to keep the search results within the scope of the SLR. In this study, the initial search was performed using a string with four keywords. The search string was (\qq{cattle} AND \qq{identification}) AND (\qq{machine learning} OR \qq{deep learning}). Some articles were extracted from the search results, and the title, abstract, and author-specified keywords were read to find the synonyms for the basic search keywords. For \qq{cattle}, synonyms considered were \qq{cow} and \qq{livestock}. For \qq{identification}, synonyms considered were \qq{recognition} and \qq{detection}. The keywords \qq{neural network}, \qq{image processing} and \qq{vision} were added with \qq{machine learning} and \qq{deep learning} as similar terms. Thus, the general search string was (\qq{cattle} OR \qq{cow*} OR \qq{livestock}) AND (\qq{identification} OR \qq{recognition} OR \qq{detection}) AND (\qq{machine learning} OR \qq{deep learning} OR \qq{neural network} OR \qq{image processing} OR \qq{vision}). The search keywords were used for articles in four databases (August 2021). The search strings for the databases are shown in Table \ref{tab:SearchKeywords}.

\begin{table}[!ht]
    \centering
    \caption{Search strings for the selected databases.}
    \label{tab:SearchKeywords}
    \begin{tabular}{p{3cm} p{11cm}}
    \hline
    \textbf{Database name} & \textbf{Search string}\\
    \hline \hline
    IEEE Xplore & ((cattle OR cow* OR livestock) AND (identification OR recognition OR detection) AND (\qq{deep learning} OR \qq{machine learning} OR \qq{neural network} OR \qq{image processing} OR vision)) (anywhere). \\
    \hline
    Science Direct & (cattle OR cow) AND (identification OR recognition OR detection) AND (\qq{deep learning} OR \qq{machine learning} OR \qq{neural network} OR \qq{image processing}). It was used to search in the title, abstract and keywords. \\
     \hline
    Scopus & TITLE-ABS-KEY ((\qq{cattle identification} OR \qq{cow* identification} OR \qq{livestock identification} OR \qq{cattle recognition} OR \qq{cow* recognition} OR \qq{livestock recognition} OR \qq{cattle detection} OR \qq{cow* detection} OR \qq{livestock detection}) AND (\qq{deep learning} OR \qq{machine learning} OR \qq{neural network} OR \qq{image processing} OR vision)). It was used to search in the title (TITLE), abstract (ABS) and keywords (KEY).\\
     \hline
     Web of Science & AB=((cattle OR cow* OR livestock) AND (identification OR recognition OR detection) AND (\qq{deep learning} OR \qq{machine learning} OR \qq{neural network})) OR AK=((cattle OR cow* OR livestock) AND (identification OR recognition OR detection) AND \qq{deep learning} OR\qq{machine learning} OR \qq{neural network})) OR TI=((cattle OR cow* OR livestock) AND (identification OR recognition OR detection) AND (\qq{deep learning} OR \qq{machine learning} OR \qq{neural network})). It was used to search in the title (TI), abstract (AB) and author keywords (AK).\\
    
    \hline
    \end{tabular}
\end{table}

This study reduced some keywords from the search string for the Science Direct database as the maximum Boolean connectors (AND/OR) for this database is eight. Since the Scopus database yielded many articles with the general search string, the search results were reduced by putting two different keywords together. In this SLR, we did not limit the publication year during the search. After performing the above search strings, a total of 731 articles were retrieved.

\subsection{Study selection criteria}
The selection criteria are used to identify the studies that can answer the research questions. In this study, inclusion and exclusion criteria were defined based on the research questions. The search results from all databases were recorded on a spreadsheet for scrutiny using the inclusion and exclusion criteria. A study was selected for the SLR when the inclusion criteria were true but the exclusion criteria were false. The exclusion criteria were: (\romannum{1}) publication is not related to ML or DL for cattle identification, (\romannum{2}) publication is a survey or review paper, (\romannum{3}) publication is not written in English. The inclusion criteria were that the publication must be applied to either the ML and/or DL approaches for cattle identification. 

\begin{figure}[ht]
    \centering
    \includegraphics[width=0.8\textwidth]{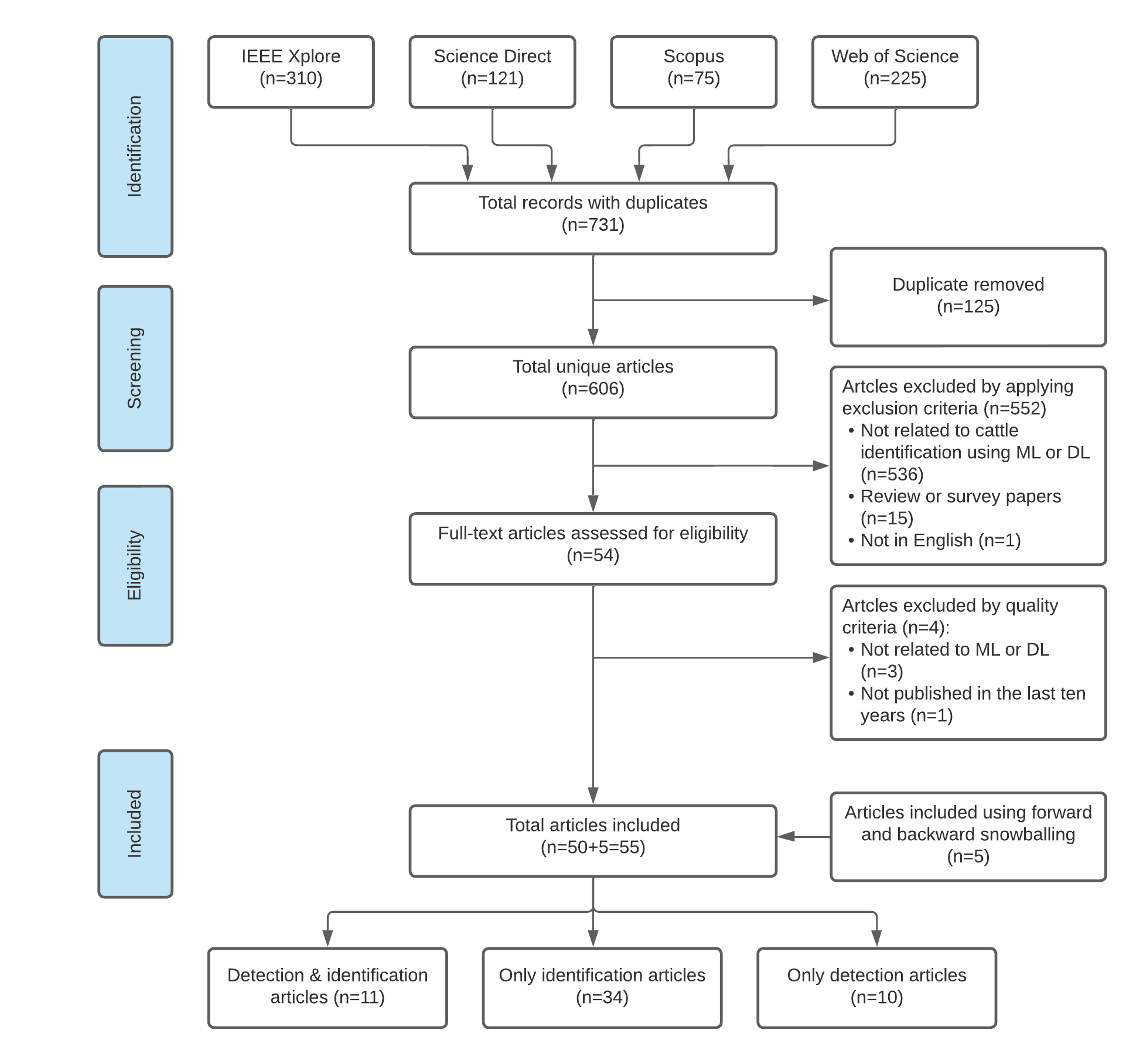}
    \caption{Article selection flowchart for the systematic literature review.}
    \label{fig:FlowChart}
\end{figure}

After excluding duplicate records (n=125), the selection criteria were applied to the rest of the records (n=606). Thus, a total of 54 full-text articles were assessed for eligibility. For selecting the final articles, we considered articles that were published in the last ten years. During the eligibility check and quality assessment, four more full-text articles were excluded because they did not satisfy the quality criteria, including not related to ML or DL for cattle identification and published more than ten years ago. The quality assessment was performed by following the study of \citet{MTDkitchenham2009systematic}. Then, this study applied backward and forward snowballing techniques \citep{MTDwohlin2014guidelines} to 50 articles and found five more relevant articles. Thus, a total of 55 articles were finally selected for this SLR. A complete flow of the article selection process is shown in Fig. \ref{fig:FlowChart} using PRISMA. We also conducted search in the Association for Computing Machinery (ACM), Springer, Inderscience and some other publishers databases with the search strings. The related articles found from the search results were duplicate to our selected databases search. For this reason, we did not report them in the PRISMA in details.

\subsection{Data extraction}
To address the research questions, each selected article was read in full and the necessary data elements were extracted from it. A spreadsheet was prepared to store all the extracted data for the selected articles. The spreadsheet columns represent the different data elements of the studies, and the rows represent the articles reviewed for the SLR. In the spreadsheet, each article is summarised by research goal, dataset, feature used, featured extraction, model, location, publishing year, performance evaluation metrics and challenges. The extracted data are then classified according to the research questions. The summarised results of this SLR are reported in Section \ref{results}.   

\section{Cattle identification overview}
\begin{figure}[htb]
    \centering
    \includegraphics[scale=.8]{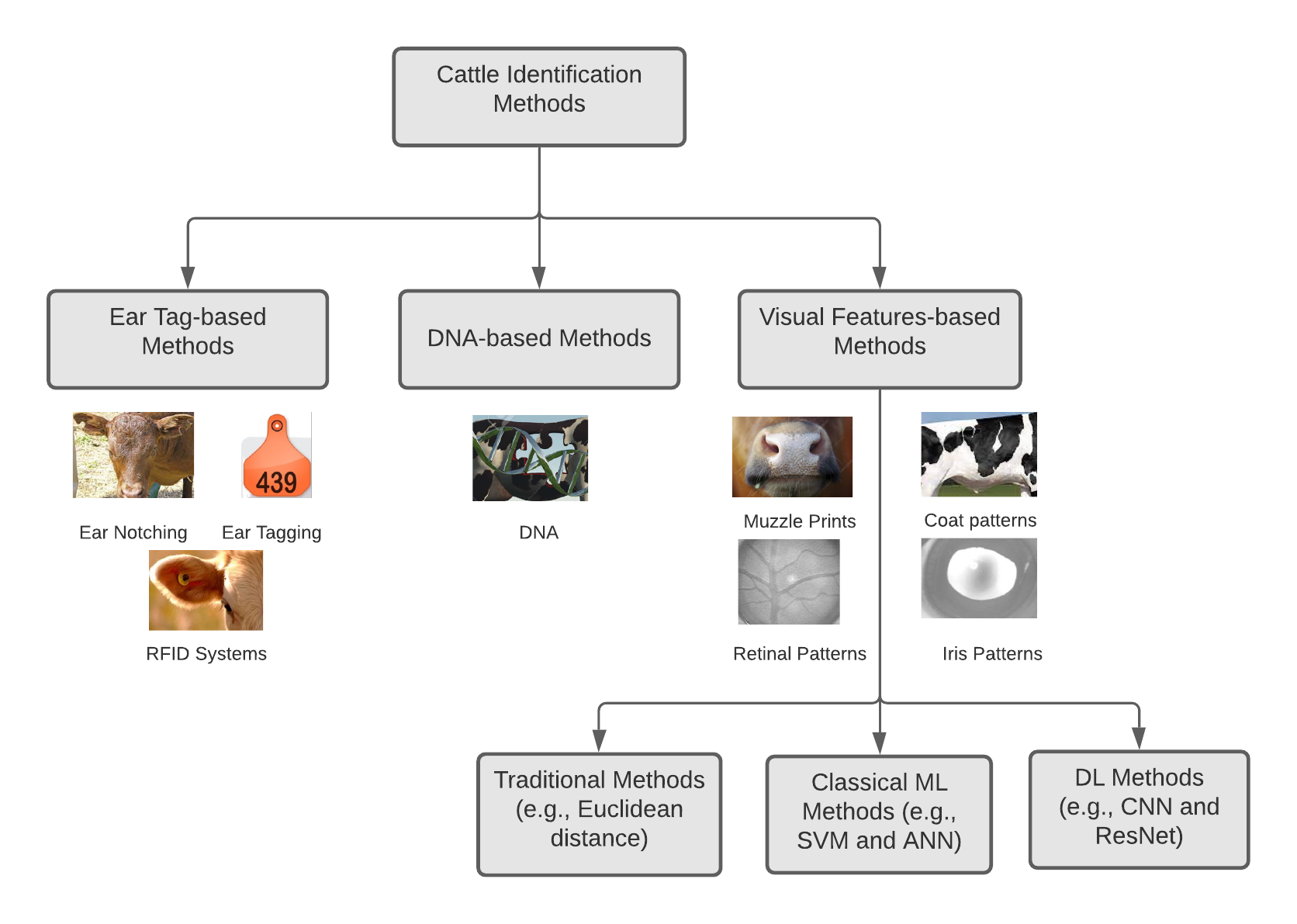}
    \caption{Different types of cattle identification methods.}
    \label{fig:ID_Overview}
\end{figure}

Cattle identification is the process of recognising cattle using unique identifiers or features \citep{LR1awad2016classical}. Accurate cattle identification plays an important role in allowing cattle farmers to implement individual animal management techniques. Individual identification also underpins genetic improvement, disease management, biosecurity, and supply chain management. Fig. \ref{fig:ID_Overview} shows the three major types of cattle identification methods -- ear tag-based methods, DNA-based methods, and visual features-based methods.

\subsection{Ear tag-based methods}
Ear tag-based cattle identification methods are widely used in livestock farm management \citep{CIOruiz2011role,CIOkumar2016fast}. These methods can help to understand disease trajectories and control the spread of acute diseases \citep{CIOvlad2012survey,CIOwang2010rfid}. Tag-based methods use unique identifiers, including permanent markings (e.g., ear notching, tattooing, and branding), temporary markings (e.g., ear tagging), and electronic devices (e.g., radio frequency identification (RFID)) \citep{LR1awad2016classical}.

Ear notching is a system of removing portions of the ear or ears of an animal so the removed portions create a distinct, recognisable shape \citep{INneary2002methods}. A combination of the positions of the ear notches is used to identify individual cattle. Ear notching has potential welfare implications, and other identification methods are more welfare friendly \citep{CIOnoonan1994behavioural}. The ear notching method applies only to a limited number of cattle on a farm. Thus, it is not used to identify individual cattle on large-sized farms.

Ear tagging is a livestock identification method \citep{LR1awad2016classical} that is widely used to provide individual cattle identification. It is quick to implement and features a low-cost identification system. An ear tag can be made of metal or plastic with bar codes, letters or numbers, and the tags can vary in size and colour. But ear tags can fall out, resulting in lost individual identification \citep{CIOwang2010rfid}. In addition, ear tags risk damage, can be duplicated, and are susceptible to fraudulent manipulation. Lost tags do not allow for the long-term recognition of individual animals \citep{CIOfosgate2006ear}. These limitations led to the development of electronic identification devices (EIDs).

Passive radio frequency identification devices (RFID) are commonly used for individual identification and tracking in livestock farming \citep{CIOruiz2011role}. The architecture of the RFID device includes an RFID tag, a communication channel, a tag reader, a server, and an RFID back-end. These devices use radio waves to transmit livestock data as a unique code made up of a sequence of numbers. However, skilled people are needed to set up and manage the RFID system. Additionally, it has some limitations in terms of security, including tag-content changes and a high possibility of system spoofing \citep{CIOroberts2006radio}.    

Although tag-based cattle identification methods have wide acceptability, well-defined research documents and long term utilisation, they have some common problems with vulnerabilities, monitoring, disease control, fraud, and cattle welfare concerns \citep{CIObowling2008identification,CIOhuhtala2007evaluation}. 

\subsection{DNA-based methods}
DNA-based cattle identification methods have been developed to identify individual cattle for the understanding of critical diseases, production management, and health monitoring. However, these methods are expensive in terms of both time and cost. DNA-based identification methods are used for specific applications, such as determining genetic linkages for genetic improvement programs. Using DNA for routine identification is not cost effective as it is a very time-consuming process to get unique DNA identifiers to identify individual animals \citep{LR2kumar2020cattle}. \citet{CIOgallinat2013dna} introduced a DNA-based identification model for bovine casein gene variants. This study used four casein genes that were sequenced in a total of 319 animals.    

\subsection{Visual features-based methods}
As the cattle identification system follows pattern recognition, it retrieves the animal's unique biometric and visual features to identify them individually. The unique features for cattle identification include the muzzle print, face, body coat pattern, and iris pattern. The biometric features-based approaches can offer an accurate and efficient solution for individual cattle identification using traditional methods (e.g., SIFT, pattern matching, and Euclidean distance), ML methods (e.g., SVM, KNN, and ANN) and DL methods (e.g., CNN, ResNet, and Inception) \citep{CIOnoviyanto2013beef,CIObarry2007using,CIOarslan20143d,AID2andrew2016automatic,AID24andrew2017visual}.

In the literature, muzzle print images have been widely used for cattle identification \citep{AID1kusakunniran2018automatic,AID5awad2019bag,AID10kumar2017muzzle,CIObarry2007using}. Like human fingerprints, the muzzle print and nose print of cattle show distinct grooves and beaded patterns. It has been recognised as a unique biometric feature since 1921 \citep{CIOpetersen1922identification}. Cattle muzzle print images can be captured using digital cameras, then feature extraction methods (e.g., SIFT and SURF) are used to extract unique features for identification.

Iris and retinal biometric features have been used for individual cattle identification. The cattle retinal pattern remains unchanged over time, and the iris contains some discriminating biometric features \citep{CIOallen2008evaluation,CIOlu2014new}. However, these methods have had limited applications due to the difficulty of capturing livestock retinal and iris images. 

The body coat pattern and face are unique features for identifying individual cattle \citep{LR2kumar2020cattle}. With the recent advent of ML and DL methods, facial and body coat patterns have been widely used for cattle identification \citep{CIOarslan20143d,AID2andrew2016automatic,AID48yang2019dairy}. For the ML models, the features are extracted from the face or body images and then fed into the models for identification. DL models have powerful feature extraction abilities for cattle identification without pre-specifying any features \citep{AID32kumar2018deep,AID24andrew2017visual}.  
 
In recent times, automatic cattle identification has gained popularity among researchers using ML and DL methods. The traditional methods cannot handle large datasets, and the accuracy of these methods is poor relative to ML and DL methods \citep{LR2kumar2020cattle}. In this SLR, ML and DL based approaches for cattle identification are considered for discussion as the other methods are out of scope in this study. 

\section{Review reports} 
\begin{figure}[htbp]
    \centering
    \includegraphics[scale=1]{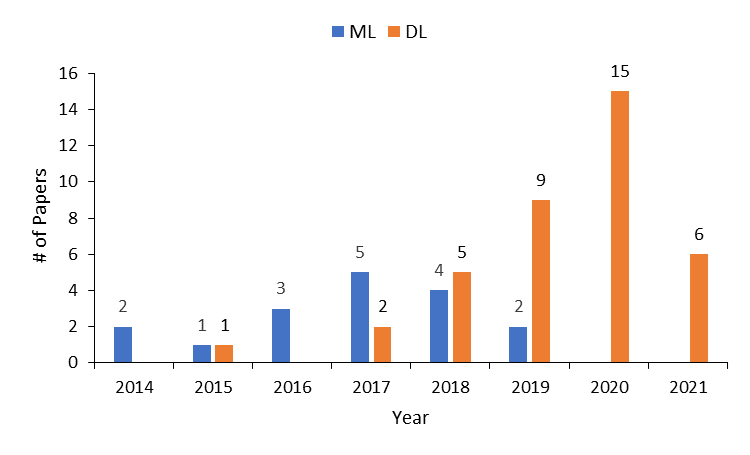}
    \caption{Distribution of selected papers in terms of year.}
    \label{fig:PublishYear}
\end{figure}

{\label{results}}
In total, 55 papers were selected for this SLR after applying the selection criteria. The year-wise distribution of these papers is shown in Fig. \ref{fig:PublishYear}. The papers for cattle identification are divided into two groups: (\romannum{1}) ML based papers and (\romannum{2}) DL based papers. The figure indicates that the number of published papers based on ML was higher than for DL based papers before 2018. This result indicates that the application of ML techniques dominated cattle identification in livestock farm management before DL approaches began gaining in popularity. Based on our survey report, the number of ML based papers for cattle identification published annually has been greater than the number of DL based papers since 2018. In addition, the figure shows that the researchers have emphasised DL models for cattle identification in recent years. This is mainly because of the higher accuracy of DL models on large datasets \citep{LR6mahmud2021systematic}. 

\begin{figure}[htb]
    \centering
    \includegraphics[width=.8\textwidth]{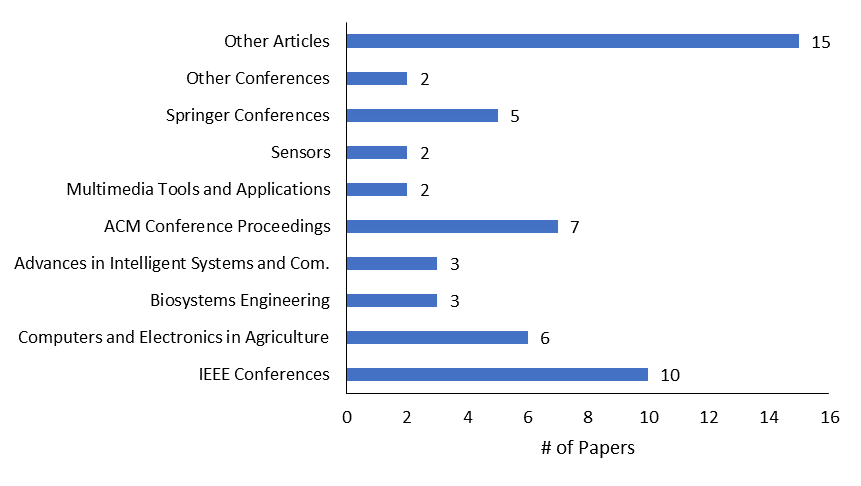}
    \caption{Distribution of journals and conferences for the selected papers related to cattle identification (published 2014-2021).}
    \label{fig:JournalName}
\end{figure}

The distribution of journals and conferences for the selected papers is presented in Fig. \ref{fig:JournalName}. The figure indicates that IEEE Conferences, ACM Conference Proceedings, and the Computer and Electronics in Agriculture journal are the three top outlets that published the highest number of automatic cattle identification papers. The other three outlets, Biosystems Engineering, Advances in Intelligent Systems and Computing, and Springer Conferences, published more than two papers in this sector.

\begin{figure}[!ht]
    \centering
    \includegraphics[width=.8\textwidth]{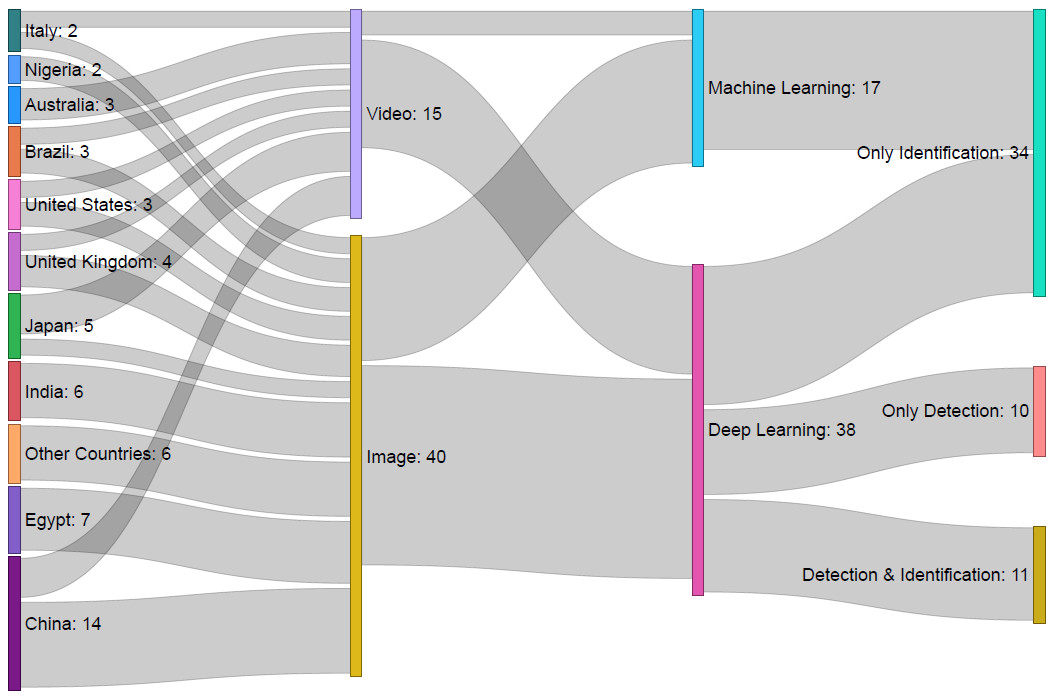}
    \caption{Country-wise summary of selected papers in terms of model, dataset type and task.}
    \label{fig:Sankeydiagram}
\end{figure}
Fig. \ref{fig:Sankeydiagram} shows a Sankey diagram for the country-wise overall summary of the selected papers in terms of dataset types and approaches used in cattle identification. This SLR reports that China published the highest number of papers (14) for cattle identification, followed by Egypt (7), India (6), and Japan (5). Most of the reviewed papers use image-based datasets (40 times) for cattle identification. Most of the video-based datasets are used in DL based papers. As this study only considers ML and DL based papers for the review, we have found 17 papers that use ML models and 38 papers that use DL models for cattle identification. The papers are divided into three groups based on the study goal(s): (\romannum{1}) only identification (34), (\romannum{2}) only detection (10), and (\romannum{3}) both detection and identification (11). The cattle identification is the process of identifying individual cattle using ML and DL approaches. In Fig. \ref{fig:Sankeydiagram}, it is noticeable that all the ML based papers are only for identification purposes, as the classical ML algorithms are only used for classification problems. The cattle detection system allows us to detect the cattle for individual identification, monitoring, and counting. It can implemented only by DL approaches.

In the following subsections we have addressed the research questions identified in Section \ref{sec:rq}.

\subsection{Machine learning approaches for cattle identification}
Classical ML approaches perform well for cattle identification in livestock farming. Table \ref{tab:ML} presents a summary of the ML based papers. The summary includes the best model, dataset, used feature, feature extraction method, and performance for automatic cattle identification. In this SLR, many ML based papers used more than one ML approach. In those cases, the performance of the best model is reported in Table \ref{tab:ML}. Based on our review, most of the ML based papers used the image dataset. The images were used to form the training and testing dataset for cattle identification. About 70\% of the papers used the cattle muzzle print as a feature because of its unique patterns. 

\begin{table}[t]
\centering
    \caption{Performance of ML models used in cattle identification research papers.}
    \label{tab:ML}
\resizebox{1\textwidth}{!}{
\begin{tabular}{lccccccc}
\hline
\multicolumn{1}{c}{\multirow{2}{*}{Reference}} & \multicolumn{1}{c}{\multirow{2}{*}{Best model}} & \multicolumn{3}{c}{Dataset}          & \multirow{2}{*}{\begin{tabular}[c]{@{}c@{}}Used \\ feature\end{tabular}} & \multirow{2}{*}{\begin{tabular}[c]{@{}c@{}}Feature extraction\\ method\end{tabular}} & \multirow{2}{*}{\begin{tabular}[c]{@{}c@{}}Best performance\\ (\%)\end{tabular}} \\ \cline{3-5}
\multicolumn{1}{c}{} & \multicolumn{1}{c}{} & Breed & Size & Split (Train,Val,Test) &&&\\
\hline \hline
       \citep{AID1kusakunniran2018automatic} & SVM & NC & 217 images & 186, --, 31 & Muzzle & LBP & 100 (A)\\
       \citep{AID2andrew2016automatic} & SVM & H & 377 images & 83, --, 294 & Body & ASIFT & 97 (A)\\
       \citep{AID3schilling2018validation} & SVM & NC & 302 images & 150, --, 152 & \makecell[t c]{Mammary\\glands} & LBP & 60 (A)\\
       \citep{AID4li2017automatic} & QDA & H & 1,965 images & 1,667, 298, -- & Tailhead & \makecell[t c]{Zernike\\moments} & 99.7 (A)\\
       \citep{AID5awad2019bag} & SVM & NC & 105 images & 75, 15, 15 & Muzzle & SURF & 93 (A)\\
       \citep{AID6zhao2019individual} & FLANN & H & 528 videos & 198, --, 330 & Body & FAST & 96.72 (A)\\
       \citep{AID7kumar2018group} & GSRC & M & 5,000 images & 3,000, --, 2000 & Muzzle & LBP & 93.87 (A)\\
       \citep{AID8lv2018image} & BruteForce & H & 1,500 images & 900, --, 600 & Body & SIFT & 98.33 (A)\\
       \citep{AID9kumar2017automatic} & ANN & M & 5,000 images & 3,000, --, 2,000 & Muzzle & LBP & 96.74 (A)\\
       \citep{AID10kumar2017muzzle} & KNN & M & 5,000 images & 3,000, --, 2,000 & Muzzle & SURF & 93.87 (A)\\
       \citep{AID11kumar2017real} & SVM & M & 5,000 images & 3,000, --, 2,000 & Muzzle & FLPP & 96.87 (A)\\
       \citep{AID12gaber2016biometric} & AdaBoost & NC & 217 images & 186, --, 31 & Muzzle & WLD & 98.9 (A)\\
       \citep{AID13zaoralek2016cattle} & SVM & NC & 322 images & --, --, -- & Muzzle & SVD & 75 (F)\\
       \citep{AID14ahmed2015muzzle} & SVM & NC & 217 images & 186, --, 31 & Muzzle & SURF & 100 (A)\\
       \citep{AID15tharwat2014cattle} & SVM & NC & 217 images & 186, --, 31 & Muzzle & Gabor filter & 99.5 (A)\\
       \citep{AID16tharwat2014cattle} & SVM & NC & 217 images & 186, --, 31 & Muzzle & LBP & 99.5 (A)\\
       \citep{AID18el2017muzzle} & ANN & NC & 1,060 images & 636, 212, 212 & Muzzle & \makecell[t c]{Box-counting} & 99.18 (A)\\

\hline

\hline
 \multicolumn{8}{l}{NC=Non-classified, H=Holstein, M=Multiple (Ongole, Punganur, Holstein, Cross and Balinese)}\\
 \multicolumn{8}{l}{A=Accuracy, F=F1 score}\\
 \multicolumn{8}{l}{QDA=Quadratic discriminant analysis, FLANN=Fast library for approximate nearest neighbors}\\
 \multicolumn{8}{l}{GSRC= Group sparsity residual constraint, FLPP=Fuzzy linear preserving projections, WLD=Weber’s native descriptor}\\
 
\end{tabular}
}
\end{table}

\begin{figure}[htb]
    \centering
    \includegraphics[scale=.8]{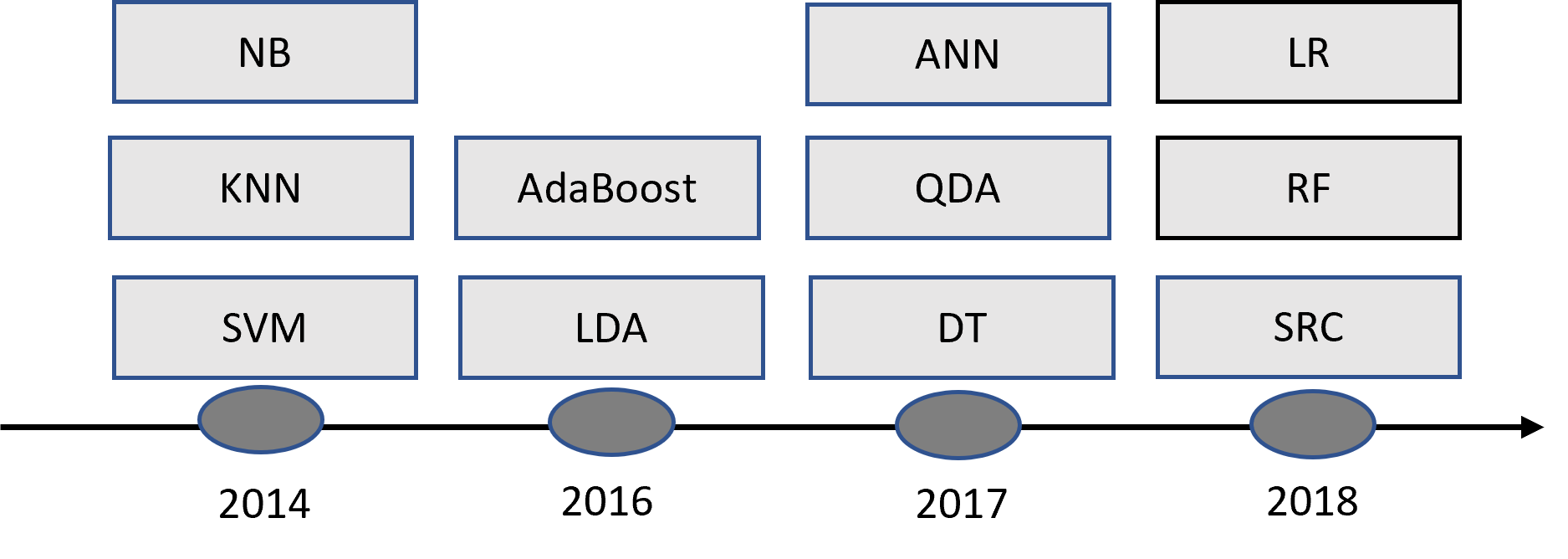}
    \caption{Timeline of the ML models used for the first time in the selected papers.}
    \label{fig:Timeline_ML}
\end{figure}

To address research question one (RQ1), ML models were analysed and listed in Table \ref{tab:TML_model}. As shown in the table, the top three most used ML models are SVM, KNN, and ANN. They are also the best models found in the most reviewed studies as shown in Table \ref{tab:ML}. Fig. \ref{fig:Timeline_ML} provides a timeline of the ML models used for the first time for cattle identification as per the reviewed papers. The figure indicates that SVM and KNN have been used since 2014 for cattle identification. ML models such as ANN and decision tree (DT) started to be used for cattle identification in 2017, whereas random forest (RF) and logistic regression (LR) started in 2018. The most used classical ML algorithms are briefly explained below.

\begin{table}[!t]
    \centering
    \caption{Traditional machine learning models used in cattle identification.}
    \label{tab:TML_model}
    \begin{tabular}{lp{11cm}c}
    \hline
       ML model & Paper reference & Count\\
     \hline\hline
     SVM & \citep{AID1kusakunniran2018automatic,AID2andrew2016automatic,AID3schilling2018validation,AID4li2017automatic,AID5awad2019bag,AID11kumar2017real,AID13zaoralek2016cattle,AID14ahmed2015muzzle,AID15tharwat2014cattle,AID16tharwat2014cattle,AID31hu2020cow,AID32kumar2018deep,AID33achour2020image} & 13\\
     
     KNN & \citep{AID3schilling2018validation,AID9kumar2017automatic,AID10kumar2017muzzle,AID12gaber2016biometric,AID16tharwat2014cattle,AID36andrew2021visual} & 6 \\
     
     ANN & \citep{AID4li2017automatic,AID6zhao2019individual,AID9kumar2017automatic,AID18el2017muzzle} & 4 \\
     
     DT & \citep{AID3schilling2018validation,AID9kumar2017automatic} & 2\\
     
     LDA & \citep{AID4li2017automatic,AID13zaoralek2016cattle} & 2 \\
     
     BruteForce & \citep{AID6zhao2019individual,AID8lv2018image} & 2\\ 
     
     Naive Bayes & \citep{AID9kumar2017automatic,AID16tharwat2014cattle} & 2 \\
     
     SRC & \citep{AID7kumar2018group} & 1 \\
     
     RF & \citep{AID3schilling2018validation} & 1 \\
     
     LR & \citep{AID3schilling2018validation} & 1 \\
     
     QDA & \citep{AID4li2017automatic} & 1 \\
     
     AdaBoost & \citep{AID12gaber2016biometric} & 1 \\
     
     Trucket decomposition & \citep{AID13zaoralek2016cattle} & 1 \\
      
      \hline
      
      \hline
      \multicolumn{2}{l}{DT=Decision tree, LDA= Linear discriminant analysis, SRC= Sparsity residual constraint}\\
     \multicolumn{2}{l}{RF=Random forest, LR= Logistic regression}\\
      
    \end{tabular}
\end{table}

SVM \citep{RRjoachims1998making} is a widely used algorithm for cattle identification. It is a technique that uses a hyperplane to classify different groups that separate the classes by maximising their marginal distance in high dimensional feature space \citep{RRjoachims1998making}. The marginal distance is the gap between the hyperplane and its nearest data points for the two classes. The dataset follows the condition that each data point belongs to only one of the two classes.

KNN \citep{RRcover1967nearest} is a simple and older machine learning classification algorithm. In KNN classification, the nearest neighbours are the data samples with minimum distance between the feature space and the new data sample. The {\lq{K}\rq} is the number of closest neighbours considered for voting to classify a new sample. A class label provided for most {\lq{K}\rq} nearest neighbours forms the training data and is defined as a predicted class for the new data sample. The different classification outcomes can generate different {\lq{K}\rq} values for the same sample example.

ANNs \citep{RRmcculloch1943logical,RRrumelhart1986learning} are machine learning algorithms developed based on the function of neural parts of the human brain. The neurons of the human brain are connected using multiple axon junctions. Thus, they form a graph-like architecture. The links among neurons help to receive, process, and store information. Similarly, an ANN algorithm can be presented as a network of interconnected nodes. According to the inter-connectivity, one node's output becomes the input of another node. A group of nodes forms a matrix, which is called a layer. An ANN can be represented by input, output, and hidden layers. The nodes and their connections have a weight that is used to adjust the signal strength which can be increased or decreased through repeated training. ANNs can classify the test data based on the training and subsequent adjustment of weights for nodes and their connections.

\subsection{Deep learning approaches for cattle identification and detection}
In recent years, deep learning (DL) has been applied successfully in livestock farm management. Various DL approaches have been proposed in this sector for different applications, including animal identification, detection, tracking, and health monitoring. Over the last few years, DL approaches have been widely used in research related to cattle identification and detection. Table \ref{tab:DL_model} presents a summary of the papers related to DL. The papers are divided into three groups: identification and detection, identification only, and detection only. The groups are generated after analysing the extracted information from the reviewed papers according to the study goal. The goal of the papers from the identification and detection category is to first detect the cattle using DL detection models and then classify the individual cattle using DL classification models. The only identification-related papers proposed automatic individual cattle identification systems using DL approaches. The only detection category papers developed cattle detection systems using DL detection algorithms. Table \ref{tab:DL_model} includes the dataset attributes, used features, feature extraction method, DL models for detection and identification, and performance of the models. It is observed that many DL based papers use cattle images as a dataset that is divided into training and testing sets. Based on our survey data, the cattle images for the head, muzzle print, and full body in terms of top and side view are widely used for detection and identification systems. Most of the DL based studies use the cattle body coat pattern as a feature because DL approaches have powerful feature extraction and image representation abilities.

{\footnotesize\tabcolsep=3pt

\begin{table}[!t]
\centering
    \caption{Performance of DL models used in cattle identification.}
    \label{tab:DL_model}
    \resizebox{1\textwidth}{!}{
\begin{tabular}{@{\extracolsep{4pt}}llccccccccc}

\hline
\multirow{2}{*}{Task} & \multirow{2}{*}{Reference} & \multicolumn{3}{c}{Dataset} & \multirow{2}{*}{\begin{tabular}[c]{@{}c@{}}Used \\ feature\end{tabular}} & \multirow{2}{*}{\begin{tabular}[c]{@{}c@{}}Feature\\extraction\end{tabular}} & \multicolumn{2}{c}{Best detection} & \multicolumn{2}{c}{Best identification} \\ \cline{3-5} \cline{8-9}\cline{10-11} & & Breed & Size & Split (Train,Val,Test) & &  & Model & \multicolumn{1}{c}{Performance (\%)} & \multicolumn{1}{c}{Model} & \multicolumn{1}{c}{Performance (\%)} \\ 
\hline \hline

    \multirow{11}{*}{\rotatebox[origin=c]{90}{Detection and identification}} & \citep{AID19chen2021angus} & An & 5,042 images & 2,521, --, 2,521 & Face \& body & -- & Mask R-CNN & -- & VGG16 & 85.4 (A)\\
    & \citep{AID20zin2020cow} & NC & 6,000 images & --, --, -- & Head & -- & YOLO & 96 (A) & CNN & 84 (A)\\
    & \citep{AID23andrew2019aerial} & H & 32 videos & 14, --, 18 & Body & \makecell[t c]{Inception v3,\\LSTM} & YOLO v2 & 92.4 (A) & LRCN & 93.6 (A)\\
    & \citep{AID24andrew2017visual} & H & \makecell[t c]{940 images\\1,064 videos} & \makecell[t c]{--, --, --\\957, --, 107} & Dorsal coat & \makecell[t c]{VGG-M 2024,\\LSTM} & \makecell[t c]{R-CNN\\} & \makecell[t c]{99.3 (mAP)\\--} & \makecell[t c]{R-CNN\\LRCN} & \makecell[t c]{96.03 (mAP)\\98.13 (A)}\\
    & \citep{AID30tassinari2021computer} & H & 11,754 frames & 10,105, 1,649, -- & Body & -- & YOLO v3 & 66 (P) & DarkNet & 73 (F)\\
    & \citep{AID31hu2020cow} & H & 958 images & 593, --, 365 & Body & Multi CNN & YOLO & -- & SVM & 98.36 (A)\\
    & \citep{AID33achour2020image} & H & 4,875 images & 2,931, --, 1,944 & Head & CNN & Xception & -- & Multi CNN & 97.06 (A)\\
    & \citep{AID36andrew2021visual} & H & 4,736 images & 1,895, 473, 2,368 & body & ResNet & YOLO v3 & 98.4 (mAP) & KNN & 93.75 (A)\\
    & \citep{AID41shen2020individual} & H & 1,433 images & 1,015, 418, -- & Body & -- & YOLO & -- & AlexNet & 96.65 (A)\\
    & \citep{AID42guan2020cattle} & NC & 1,650 images & 1,100, 550, -- & Face \& body & -- & RefineDet & 87.4 (A) & CNN & 80 (A)\\
    & \citep{AID47yao2019cow} & NC & 18,231 images & 14,585, --, 3,646 & Face & ResNet101 & Faster R-CNN & 99.6 (A) & PnasNet-5 & 94.7 (A)\\
    \hline
    \multirow{17}{*}{\rotatebox[origin=c]{90}{Only identification}} & \citep{AID21phyo2018hybrid} & NC & 13,603 frames & 9,064, --, 4,539 & Body & Skew histogram & -- & -- & 3D-CNN & 96.3 (A)\\
    & \citep{AID22qiao2020bilstm} & NC & 363 videos & 288, --, 75 & Body & BiLSTM & -- & -- & Custom & 91 (A)\\
    & \citep{AID25manoj2021identification} & NC & 150 images & --, --, -- & Body & SIFT & -- & -- & CNN & --\\
    & \citep{AID26bergamini2018multi} & NC & 17,802 images & 12,952, 4,289, 561 & Body & CNN & -- & -- & 3D-CNN & 89.1 (A)\\
    & \citep{AID28yukun2019automatic} & H & 3,430 images & 2,400, --, 1,030 & Body & CNN & -- & -- & DenseNet & 98.5 (A)\\
    & \citep{AID29santoni2015cattle} & M &  775 images &  620, --, 155 & Body & GLCM & -- & -- & CNN and LeNet-5 & 98.92 (A)\\
    & \citep{AID32kumar2018deep} & NC & 5,000 images & 4,000, --, 1,000 & Muzzle & CNN & -- & -- & DBN & 98.99 (A)\\
    & \citep{AID34qiao2019individual} & NC & 516 videos & 439, --, 77 & Body & LSTM & -- & -- & Custom & 91 (A)\\
    & \citep{AID35de2020recognition} & P & 27,849 images & 25,085, --, 2,764 & Body & CNN & -- & -- & DenseNet201 & 99.86 (A) (A)\\
    & \citep{AID39wang2020method} & H & 2880 activity & 2,304, 576, -- & Activity & Min \& Max & -- & -- & DNN & 93.81 (A)\\
    & \citep{AID44bello2020deep} & NC & 4,000 images & 3,000, --, 1,000 & Nose & CNN & -- & -- & DBN & 98.99 (A)\\
    & \citep{AID45bello2020image} & NC & 1,000 images & 400, --, 600 & Body & CNN & -- & -- & SDAE & 89.95 (A)\\
    & \citep{AID48yang2019dairy} & H & 85,200 images & 82,010, --, 3,190 & Face & CNN & -- & -- & CNN and ResNet50 & 94.92 (A)\\
    & \citep{AID50zin2018image} & H & 22 videos & --, --, -- & Body & -- & -- & -- & 3D-CNN & 97.01 (A)\\
    & \citep{AID51li2018cow} & NC & 21,600 images & 19,440,--, 2,160 & Body & CNN & -- & -- & Inception v3 & 98 (A)\\
    & \citep{AID53bhole2019computer} & H & 1,237 images & --, --, -- & Body & CNN & -- & -- & AlexNet & 97.5 (A)\\
    & \citep{AID54wang2020cattle} & S & 187 images & 65, --, 122 & Face & CNN & -- & -- & VGG16 & 93 (A)\\
    \hline
    \multirow{10}{*}{\rotatebox[origin=c]{90}{Only detection}} & \citep{AID27xu2020automated} & NC & 750 images & 500, --, 250 & Body & CNN & Mask R-CNN & 94 (A) & -- & --\\
    & \citep{AID37wang2020mtfcn} & NC & 1,323 images & 926, --, 397 & Face & -- & BaseNet & 95 (P) & -- & --\\
    & \citep{AID38barbedo2020cattle} & NC & 15,410 images & 12,328, --, 3,082 & Body & -- & Xception & 87 (A) & -- & --\\
    & \citep{AID40zuo2020livestock} & NC &3,139 images & 2,825, --, 314 & Body &  -- & MultiResUNet & 95.37 (A) & -- & --\\
    & \citep{AID43shao2020cattle} & NC & 656 images & 245, --, 411 & Body & -- & YOLO v2 & 95.7 (P) & -- & --\\
    & \citep{AID46han2019livestock} & NC & 43 images & 36, --, 7 & Body & -- & Faster R-CNN & 89.1 (P) & -- & --\\
    & \citep{AID49rivas2018detection} & NC & 13,520 images & 10,816, --, 2,704 & Body & -- & CNN & 95.5 (A) & -- & --\\
    & \citep{AID52barbedo2019study} & C & 17,258 images & 13,806, --, 3,452 & Body & CNN & NasNet large & 99.2 (A) & -- & --\\
    & \citep{AID55lin2019object} & NC & 1,000 images & --, --, -- & Body & SURF & Fast R-CNN & 96.9 (A) & -- & --\\
    & \citep{AID56aburasain2020drone} & NC & 300 images & 270, --, 30 & Body & -- & YOLO v3 & 100 (P) & -- & --\\

\hline
    
\hline
 \multicolumn{11}{l}{NC=Non-classified, An=Angus, H=Holstein, P=Pantaneira, M=Multiple (Ongole, Punganur, Holstein, Cross and Balinese), S=Simmental, C=Canchim}\\
 \multicolumn{11}{l}{A=Accuracy, F=F1 score, P= Precision, mAP= Mean average precision}\\
 \multicolumn{11}{l}{LSTM= Long short-term memory, BiLSTM= Bidirectional long short-term memory, DNN= Deep neural network, DBN= Deep belief network}\\
 \multicolumn{11}{l}{SDAE= Stacked denoising autoencoder, GLCM= Gray level co-occurrence matrices}
 
\end{tabular}
}
\end{table}
}

\begin{table}[!t]
    \centering
    \caption{Deep learning models used in cattle identification.}
    \label{tab:DL_Identification_model}
    \begin{tabular}{lp{11cm}c}
    \hline
       Identification model & Paper reference & Count\\
     \hline\hline
      CNN & \citep{AID20zin2020cow,AID25manoj2021identification,AID29santoni2015cattle,AID33achour2020image,AID42guan2020cattle,AID44bello2020deep,AID48yang2019dairy} & 7\\
      ResNet & \citep{AID19chen2021angus,AID35de2020recognition,AID41shen2020individual,AID47yao2019cow,AID48yang2019dairy,AID52barbedo2019study}	& 6\\
      Inception & \citep{AID35de2020recognition,AID40zuo2020livestock,AID46han2019livestock,AID51li2018cow,AID52barbedo2019study} & 5\\
      VGG & \citep{AID19chen2021angus,AID41shen2020individual,AID47yao2019cow,AID52barbedo2019study,AID54wang2020cattle} & 5\\
      DenseNet & \citep{AID28yukun2019automatic,AID35de2020recognition,AID41shen2020individual,AID52barbedo2019study}	& 4\\
      DCNN & \citep{AID21phyo2018hybrid,AID26bergamini2018multi,AID50zin2018image} & 3\\
      AlexNet & \citep{AID41shen2020individual,AID47yao2019cow,AID53bhole2019computer} & 3\\
      DBN & \citep{AID32kumar2018deep,AID44bello2020deep,AID45bello2020image} & 3\\
      SDAE & \citep{AID32kumar2018deep,AID44bello2020deep,AID46han2019livestock} & 3\\
      LRCN & \citep{AID23andrew2019aerial,AID24andrew2017visual}	& 2\\
      NasNet & \citep{AID47yao2019cow,AID52barbedo2019study} & 2\\
      LeNet & \citep{AID29santoni2015cattle,AID47yao2019cow} & 2\\
      PrimNet & \citep{AID19chen2021angus} & 1\\
      R-CNN & \citep{AID24andrew2017visual} & 1\\
      DarkNet & \citep{AID30tassinari2021computer} & 1\\
      DNN & \citep{AID39wang2020method} & 1\\

      \hline
      
      \hline
      \multicolumn{3}{l}{LRCN= Long-term recurrent convolutional network}
      
    \end{tabular}
\end{table}

\begin{figure}[!t]
    \centering
    \includegraphics[scale=1]{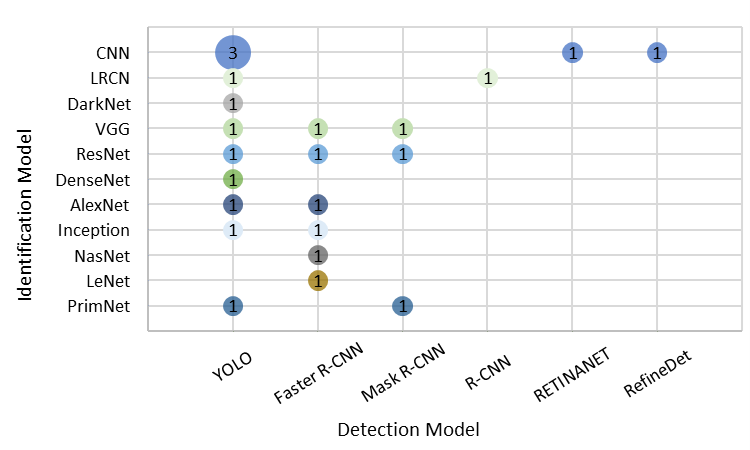}
    \caption{Combination of detection and identification models used for cattle identification.}
    \label{fig:DetectionIdentification}
\end{figure}

\begin{figure}[!t]
    \centering
    \includegraphics[scale=1]{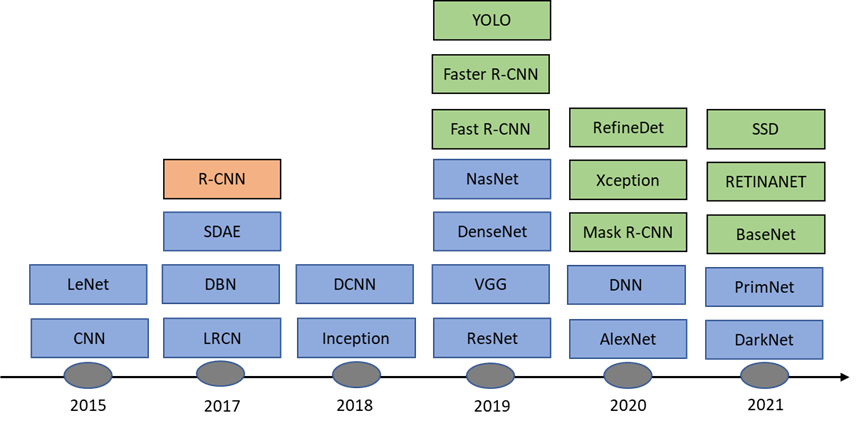}
    \caption{Timeline of when use of the DL identification and detection models were first reported to the selected papers. Blue: Identification model, green: detection model and orange: detection and identification model.}
    \label{fig:Timeline_DL}
\end{figure}

To address the second research question (RQ2), we explored and summarised DL detection and identification models. Table \ref{tab:DL_Identification_model} shows a list of DL models used in the DL based papers for cattle identification. The top three identification models are the convolutional neural network (CNN), residual network (ResNet) and Inception models. Cattle detection models are listed in Table \ref{tab:detection_model}. You Only Look Once (YOLO) and Faster R-CNN are the top two detection models used in the reviewed papers. In Fig. \ref{fig:DetectionIdentification}, we have created a matrix plot that shows the combination of detection and identification models used for cattle identification. The result shows that the YOLO detection model is used the most with the CNN identification model. It is also observed that ResNet and VGG are used as identification models where they are combined with three different detection models (i.e., YOLO, Faster R-CNN and Mask R-CNN). Fig. \ref{fig:Timeline_DL} shows a timeline of the DL identification and detection models that were used for the first time for cattle identification in the reviewed papers. It is observed that DL based classification models were first used for cattle identification in 2015, whereas the detection models were used from 2019. Based on our survey, CNN and LeNet are the first DL models used for cattle identification, as they are the oldest models in DL techniques. Short descriptions of the frequently used DL models are given below.

\begin{table}[!t]
    \centering
    \caption{Deep learning detection models used in cattle identification.}
    \label{tab:detection_model}
    \begin{tabular}{lp{11cm}c}
    \hline
       Detection model & Paper reference & Count\\
     \hline\hline
      YOLO & \citep{AID20zin2020cow,AID23andrew2019aerial,AID30tassinari2021computer,AID31hu2020cow,AID36andrew2021visual,AID41shen2020individual,AID43shao2020cattle,AID46han2019livestock,AID56aburasain2020drone} & 9\\
      Faster R-CNN & \citep{AID36andrew2021visual,AID46han2019livestock,AID47yao2019cow} & 3\\
      Mask R-CNN & \citep{AID19chen2021angus,AID27xu2020automated} & 2\\
      Xception & \citep{AID33achour2020image,AID38barbedo2020cattle} & 2\\
      R-CNN & \citep{AID24andrew2017visual} & 1\\
      RETINANET & \citep{AID36andrew2021visual} & 1\\
      BaseNet & \citep{AID37wang2020mtfcn} & 1\\
      RefineDet & \citep{AID42guan2020cattle} & 1\\
      Fast R-CNN & \citep{AID55lin2019object} & 1\\
      SSD & \citep{AID56aburasain2020drone} & 1\\
      \hline
      
      \hline
      \multicolumn{3}{l}{SSD= Single-shot detector}
    \end{tabular}
\end{table}

CNN \citep{DISalzubaidi2021review} is one of the most popular models in the field of DL. It is a widely used model for cattle identification, perhaps because it is the oldest DL model. Another explanation could be that the CNN model is simpler to explain and apply than the other DL models. The CNN architecture comprises three layers: convolutional, pooling, and fully connected layers. Convolutional layers are the significant parts of the CNN model. They consist of convolutional filters (or kernels) and feature maps. Each filter has weighted inputs, and they are convolved with the input volume to create a feature map. Each convolutional filter can create one feature map \citep{RRbrownlee2019deep}. The function of the pooling layer is to shrink the larger feature maps into smaller feature maps. It is also used to reduce overfitting. Fully connected layers are placed at the end of the CNN model and are used as the CNN classifier. Thus, the output of the fully connected layers represents the final model output. 

ResNet \citep{RRhe2016deep} is one of the most used DL models in cattle identification (see Table \ref{tab:DL_Identification_model}). It was developed to solve the vanishing and exploding gradients problem. During the training time of the deep neural network (DNN) with back-propagation, the hidden layers and derivatives are multiplied by each other. The vanishing gradients problem occurs when the gradients rapidly decrease or even vanish due to the smaller value of derivatives. The exploding gradients problem occurs when the gradients rapidly increase due to the larger value of derivatives. To avoid the vanishing and exploding gradients problem, ResNet uses the skip connections technique to skip one or more layers and connect them to the output layer. The DL network can learn from residual mapping instead of underline mapping using the skip connections technique.

The Inception model \citep{RRszegedy2015going} is a state-of-the-art DL model used for classification and detection problems. In cattle farm management, it has been widely used for cattle identification and detection. An inception network is a combination of repeating components referred to as inception modules. Each module consists of the input layer, a 1x1 convolution layer, a 3x3 convolution layer, a 5x5 convolution layer, a max pooling layer, and concatenation layer. The 1x1 convolution layer reduces the input dimensions, and it can learn patterns across the input depth. The 3x3 and 5x5 convolution layers can learn patterns across all input dimensions. The max pooling layer reduces the input dimensions to create a smaller output. The outputs of the convolution layer and the max pooling layer are concatenated in the concatenation layer.

This SLR found that YOLO \citep{RRredmon2016you} is the most used detection algorithm for cattle detection problems. It is a popular and well-performing object detection model. The YOLO algorithm follows the regression model, and it predicts classes and bounding boxes for the full object in one run of the algorithm. It has several versions, including YOLO, YOLO-V2, and YOLO-V3. The localisation accuracy was improved in YOLO-V2 by adding batch normalisation to the convolutional layers. Additionally, it increased the image resolution and used anchor boxes to predict bounding boxes \citep{RRredmon2017yolo9000}. In YOLO-V3 \citep{RRredmon2018yolov3}, the authors increased the convolutional layers to 106 and built residual blocks and applied the skip connections technique to improve the object detection performance. 

The region-based CNN (R-CNN) \citep{RRgirshick2014rich} is another powerful object detection model that is part of the family of CNN models. R-CNN has improved versions named Fast R-CNN, Faster R-CNN and Mask R-CNN. Faster R-CNN consists of two modules \citep{RRren2015faster}. The first module, called a fully convolutional network, proposes regions, whereas the second module, named the Fast-RCNN detector, uses the proposed regions. These two modules together form a single and unified DL network for object detection. This study found that the Faster-RCNN is one of the most used detection algorithms for cattle detection.

\subsection{Datasets}
A dataset plays an important role in achieving the best performance in any study. To address the third research question (RQ3), we have summarised the cattle datasets used in the reviewed articles as shown in Table \ref{tab:datasets}. The datasets are summarised in terms of several factors, including breed, the number of cattle, data type and size, image resolution, capture location, and acquisition device. The table indicates that many reviewed papers used the dataset for the Holstein cattle breed. This is mainly because of the unique coat patterns and patches on Holstein cattle bodies. The cattle coat patterns are unique features in the ML and DL approaches for identification. Our SLR reported that the most used dataset type was the image-based dataset (40 times), which included different data acquisition systems such as RGB (red, green, and blue), grey, and near-infrared (NIR) images. The images were captured from different positions of the cattle, including the side, top, rear, and front. A total of 15 studies used a video-based dataset. Of these, most of the data were collected using a digital camera. A few studies used unmanned aerial vehicles (UAVs)/drones as data acquisition devices. The cattle images or videos were captured from the cattle farm (indoor or outdoor), and then processed to develop a decision-making system using ML or DL for cattle identification. In the reviewed articles, different image resolutions are found in the datasets. Image resolution indicates the number of pixels in an image. A high resolution (i.e., a higher number of pixels) ensures the higher quality of cattle images, although it requires more storage space due to the large file size. It should be noted that high-resolution images are not necessary for the highest accuracy. Most of the reviewed studies used lower resolution (i.e., a lower number of pixels) cattle images as inputs to the ML or DL models and achieved good results for cattle identification.   
   
{\footnotesize\tabcolsep=.8pt
\begin{center}
    \begin{longtable}{p{3cm}ccccccllc}
    \caption{Datasets used in the literature.}
    \label{tab:datasets}\\
    \hline
      Reference & Breed & \# of animals & Type & Data size & Resolution & Capture location & Acquisition device & Location & Link \\
      \hline\hline
      \endfirsthead
      
      \multicolumn{10}{c}%
      {\tablename\ \thetable{} -- Continued from previous page} \\
      \hline
        Reference & Breed & \# of animals & Type & Data size & Resolution & Capture location & Acquisition device & Location & Link \\
        \hline \hline
        
    \endhead

    \hline \multicolumn{10}{r}{{Continued on next page}} \\ 
     \endfoot

     \hline
     \endlastfoot

      \citep{AID1kusakunniran2018automatic,AID12gaber2016biometric,AID14ahmed2015muzzle,AID15tharwat2014cattle,AID16tharwat2014cattle} & NC & 31 & Image (M) & 217 & 300$\times$400 & -- & Camera & Egypt & --\\
      \citep{AID2andrew2016automatic} & H & 50 & Image (B) & 377 & -- & Indoor & Kinect 2 & UK & \romannum{1}\\
      \citep{AID3schilling2018validation} & NC & 151 & Image (M.G) & 302 & 3840$\times$2160 & Indoor & Camera & US & --\\
      \citep{AID4li2017automatic} & H & 10 & Image (T) & 1,965 & 1920$\times$1080 & Indoor & DS-2CD2T32(D)-I3 & China & --\\
      \citep{AID5awad2019bag} & NC & 15 & Image (M) & 105 & -- & -- & Camera & Egypt & --\\
      \citep{AID6zhao2019individual} & H & 66 & Video (B) & 528 & 1280$\times$720  & Outdoor & Nikon D5200 & US & --\\
      \citep{AID7kumar2018group,AID9kumar2017automatic,AID10kumar2017muzzle,AID11kumar2017real,AID32kumar2018deep} & M & 500 & Image (M) & 5,000 & 500$\times$500 & Indoor & Camera & India & --\\
      \citep{AID8lv2018image} & H & 60 & Video (B) & 1,500 images & 1280$\times$720 & Outdoor & Camera & China & --\\
      \citep{AID13zaoralek2016cattle} & NC & 46 & Image (M) & 322 & 300$\times$400 & -- & Camera & CZ & --\\
      \citep{AID18el2017muzzle} & NC & 53 & Image (M) & 1,060 & -- & -- & Camera & Egypt & --\\
      \citep{AID19chen2021angus} & An & 216 & Image (F \& B) & 5,042 & -- & Outdoor & Camera & US & --\\
      \citep{AID20zin2020cow} & NC & -- & video (He) & 6,000 images & -- & Indoor & AXIS P1448-LE & Japan & --\\
      \citep{AID21phyo2018hybrid} & NC & 60 & Video (B) & 13,603 images & -- & Indoor & Camera & Japan & --\\
      \citep{AID22qiao2020bilstm} & NC & 50 & Video (B) & 36 & 401$\times$506 & Indoor & ZED camera &	Australia & --\\
     \citep{AID23andrew2019aerial} & H & 17 & Video (B) & 32 & 720$\times$720 & Outdoor & DJI Matrice drone & UK & --\\
     \citep{AID24andrew2017visual} & H & 112 & \makecell[t c]{Image (B)\\Video (B)} & \makecell[t c]{940\\1,064 videos} & \makecell[t c]{--\\3840$\times$2160} & \makecell[t c]{Indoor \\ Outdoor} & \makecell[t l]{Kinect 2\\ DJI Inspire drone} & UK & \romannum{2}\\
     \citep{AID25manoj2021identification} & NC & 26 & Image (B) & 150 & 640$\times$480 & Outdoor & Camera &	India & --\\
     \citep{AID26bergamini2018multi} & NC & 439 & Image (B) &	17,802 & -- & Indoor & Camera & Italy	 & --\\
     \citep{AID27xu2020automated} & NC & -- & Video (B) & 750 images & 512$\times$512 & Outdoor & MAVIC Pro drone & Australia & --\\
     \citep{AID28yukun2019automatic} & H & 686 & Back image (B) & 3,430 & -- & Indoor & Camera & China & --\\
     \citep{AID29santoni2015cattle} & M & 5 & Image (B) & 775 & -- & Outdoor & Camera &	Indonesia &	--\\
     \citep{AID30tassinari2021computer} & H & 4 & Video (B) & 11,745 frames & -- & Indoor & HDR-CX115E & Italy & --\\
     \citep{AID31hu2020cow} & H & 93 & Side-view image (B) & 958 & 640$\times$480 & Indoor & ASUS Xtion2 & China &	--\\
     \citep{AID33achour2020image} & H & 17 & Top-view image (He) & 4,875 & 640$\times$480 & Indoor & Yudanny Webcam & Algeria & --\\
     \citep{AID34qiao2019individual} & NC & 41 & Rear-view video (B) & 516 & 401$\times$506 & Indoor & ZED camera & Australia & --\\
     \citep{AID35de2020recognition} & P & 51 & Video (B) & 212 & -- & Outdoor & Video Recorder &	Brazil & --\\
     \citep{AID36andrew2021visual} & H & 46 & Top-view image (B) & 4,736 & -- & \makecell[t l]{Indoor \& \\ Outdoor} & \makecell[t l]{DJI Inspire MkI,\\Kinect 2} & UK & \romannum{3}\\
     \citep{AID37wang2020mtfcn} & NC & -- & Image (F) & 1,323 & -- & Indoor & Camera &	China & --\\
     \citep{AID38barbedo2020cattle} & NC & -- & Image (B) & 15,410 & -- & Outdoor & DJIMavic 2 Pro & Brazil & --\\
     \citep{AID39wang2020method} & H & 5 & Activity (B) & 14,400 & -- & Outdoor & -- & China & --\\
     \citep{AID40zuo2020livestock} & NC & -- & Image (B) & 3,139 & 3000$\times$4000 & Outdoor & Quadcopter & China & \romannum{4}\\
     \citep{AID41shen2020individual} & H & 105 & Side-view image (B) & 1,433 & 640$\times$480 & Indoor & ASUS Xtion2 & China & --\\
     \citep{AID42guan2020cattle} & NC & -- & Video (F \& B) & 1,650 frames & -- & Indoor & Camera & Japan & --\\
     \citep{AID43shao2020cattle}	& NC & 212 & Image (B) & 656 & 2122$\times$2122 & Outdoor & DJI Phantom drone &	Japan & --\\
     
     \citep{AID44bello2020deep} & NC & 400 & Image (M) & 4,000 & -- & Indoor & Camera & Nigeria & --\\
     \citep{AID45bello2020image} & NC &	10 species & Side-view image (B) & 1,000 & -- & Outdoor & CCD camera & Nigeria & --\\
     \citep{AID46han2019livestock} &	NC & -- & Image (B) & 43 & 4000$\times$4000 & Outdoor & Quadcopter & China & \romannum{4}\\
     \citep{AID47yao2019cow} & NC & 200 & Image (F) & 18,231 & -- & Indoor & Camera	& China & --\\
     \citep{AID48yang2019dairy} & H & 1,000 & Image (F) & 85,200 & -- & Indoor & Camera & China & --\\
     \citep{AID49rivas2018detection} & NC & -- & Image (B) & 13,520 & -- & Outdoor & Multirotors drone &	Spain & --\\
     \citep{AID50zin2018image} & H & 45 & Video (B) & 15 & 840$\times$400 & Indoor & Camera & Japan & --\\
     
     \citep{AID51li2018cow} & NC & 30 & Video (B) & 21,600 & -- & Outdoor & Camera & China & --\\
     \citep{AID52barbedo2019study} &	C & -- & Image (B) & 1,853 & -- & Outdoor & JI Phantom 4 Pro & Brazil & --\\
     \citep{AID53bhole2019computer} & H & 136 & Image (B) & 1,237 & 640$\times$480 & Indoor & FLIR E6 & Netherlands & --\\
     \citep{AID54wang2020cattle} & S & 36 & Video (F) & 187 Images & -- & Indoor & Camera & China & --\\
     \citep{AID55lin2019object} & NC & -- & Image (B) & 1,000 & 866$\times$652 & Outdoor & Camera &	China & --\\
     \citep{AID56aburasain2020drone} & NC & -- & Image (B) & 300 & 608$\times$608 & Outdoor & Drone & UAE & --\\

      \hline
      
      \hline
      \multicolumn{10}{l}{Breed: NC=Non-classified, An=Angus, H=Holstein, P=Pantaneira, M=Multiple, S=Simmental, C=Canchim}\\
      \multicolumn{10}{l}{Type: M=Muzzle, B=Body, T=Tailhead, M.G=Mammary glands, F=Face, He=Head}\\
      \multicolumn{10}{l}{Location: UK= United Kingdom, US= United States, UAE= United Arab Emirates, CZ=Czech Republic}\\
      \multicolumn{10}{l}{$^{\romannum{1}}$FriesianCattle2015 Dataset -- \url{http://data.bris.ac.uk}}\\
      \multicolumn{10}{l}{$^{\romannum{2}}$FriesianCattle2017 and AerialCattle2017 Dataset -- \url{http://data.bris.ac.uk}}\\
      \multicolumn{10}{l}{$^{\romannum{3}}$OpenCows2020 Dataset -- \url{http://data.bris.ac.uk}}\\
      \multicolumn{10}{l}{$^{\romannum{4}}$\url{https://github.com/hanl2010/Aerial-livestock-dataset/releases}}\\
    \end{longtable}
\end{center}
}

\subsection{Feature extraction methods}
A cattle identification dataset stores the cattle images or frames captured from videos. For individual cattle identification, the features of the cattle images are extracted using feature extraction methods and stored in the feature dataset along with a specific cattle number. The features are then fed into a DL or ML model for training and testing.
\begin{table}[hp]
    \centering
    \caption{Feature extraction (FE) methods used with ML and DL in cattle identification.}
    \label{tab:FeatureExtractor}
    \resizebox{.9\textwidth}{!}{
    \begin{tabular}{l p{11cm}c}
    \hline
     FE method & Paper reference & Count\\
     \hline\hline
     CNN & \citep{AID26bergamini2018multi} \citep{AID27xu2020automated} \citep{AID28yukun2019automatic} \citep{AID31hu2020cow} \citep{AID32kumar2018deep} \citep{AID33achour2020image} \citep{AID35de2020recognition} \citep{AID44bello2020deep} \citep{AID45bello2020image} \citep{AID49rivas2018detection} \citep{AID51li2018cow} \citep{AID52barbedo2019study} \citep{AID53bhole2019computer} \citep{AID54wang2020cattle} \citep{AID15tharwat2014cattle} & 15\\
     LBP & \citep{AID1kusakunniran2018automatic} \citep{AID3schilling2018validation} \citep{AID7kumar2018group} \citep{AID9kumar2017automatic} \citep{AID10kumar2017muzzle} \citep{AID16tharwat2014cattle} & 6\\
     SURF & \citep{AID5awad2019bag} \citep{AID6zhao2019individual} \citep{AID7kumar2018group} \citep{AID10kumar2017muzzle} \citep{AID14ahmed2015muzzle} \citep{AID55lin2019object} & 6\\
     SIFT & \citep{AID2andrew2016automatic} \citep{AID6zhao2019individual} \citep{AID7kumar2018group} \citep{AID8lv2018image} \citep{AID25manoj2021identification} & 5\\
     Inception & \citep{AID22qiao2020bilstm} \citep{AID23andrew2019aerial} \citep{AID24andrew2017visual} \citep{AID34qiao2019individual} \citep{AID47yao2019cow} & 5\\
     LSTM & \citep{AID22qiao2020bilstm} \citep{AID23andrew2019aerial} \citep{AID24andrew2017visual} \citep{AID34qiao2019individual} & 4\\
     VGG & \citep{AID24andrew2017visual} \citep{AID47yao2019cow} & 2 \\
     ResNet & \citep{AID36andrew2021visual} \citep{AID47yao2019cow} & 2\\
     Zermike moments & \citep{AID4li2017automatic} & 1\\
     MSER & \citep{AID5awad2019bag} & 1\\
     FAST & \citep{AID6zhao2019individual} & 1\\
     ORB & \citep{AID6zhao2019individual} & 1\\
     HOG & \citep{AID9kumar2017automatic} & 1\\
     LTE & \citep{AID9kumar2017automatic} & 1\\
     WLD & \citep{AID12gaber2016biometric} & 1\\
     SVD & \citep{AID13zaoralek2016cattle} & 1\\
     FLPP & \citep{AID11kumar2017real} & 1\\
     Gabor filter & \citep{AID15tharwat2014cattle} & 1\\
     Box-counting & \citep{AID18el2017muzzle} & 1\\
     GLCM & \citep{AID29santoni2015cattle} & 1\\
     DBN & \citep{AID32kumar2018deep} & 1\\
     SDAE & \citep{AID32kumar2018deep} & 1\\

      \hline
      
      \hline
      \multicolumn{3}{l}{MSER= Maximally stable extremal regions, FAST= Features from accelerated segment test}\\
     \multicolumn{3}{l}{ORB= Oriented FAST and rotated BRIEF, HOG= Histogram oriented gradient}\\
      \multicolumn{3}{l}{LTE= Laws texture energy, SVD= Singular value decomposition}
      
    \end{tabular}
    }
\end{table}

To address the fourth research question (RQ4), feature extraction methods are investigated and summarised. Table~\ref{tab:FeatureExtractor} shows the list of different feature extraction methods used in the ML and DL models for cattle identification. The top five most used feature extraction methods are CNN, local binary pattern (LBP), speeded up robust features (SURF), scale-invariant feature transform (SIFT), and Inception. The other feature extraction methods are LSTM, VGG, FAST, HOG, ORB, and LTE. Short descriptions of the most used feature extraction methods are given below. 

LBP is a feature extraction method used in the ML approaches \citep{AID3schilling2018validation,AID6zhao2019individual}. It is a texture spectrum model that assigns a binary value to each pixel in the image using the threshold of the nearest pixels around it. When the value of the nearest pixel is equal to or higher than the threshold value, it is set to 1. Otherwise, the LBP value of that pixel is set to 0. After assigning the binary values to all pixels, they are converted into decimal numbers. 

SIFT is a popular and widely used algorithm to detect local features. The algorithm seeks features by looking at interesting points in an image as well as descriptors related to scale and orientation. Thus, it achieves good outcomes in matching image feature points \citep{RRlowe1999object}. 

SURF is another powerful algorithm to detect and describe local feature points of an image \citep{RRbay2006surf}. Like SIFT, it performs three tasks, including the extraction of feature points, descriptions of feature points, and matching feature points. However, the SURF algorithm was later updated for high-performance efficiency \citep{RRbay2008speeded}.

\begin{figure}[t]
    \centering
    \includegraphics[width=.8\textwidth]{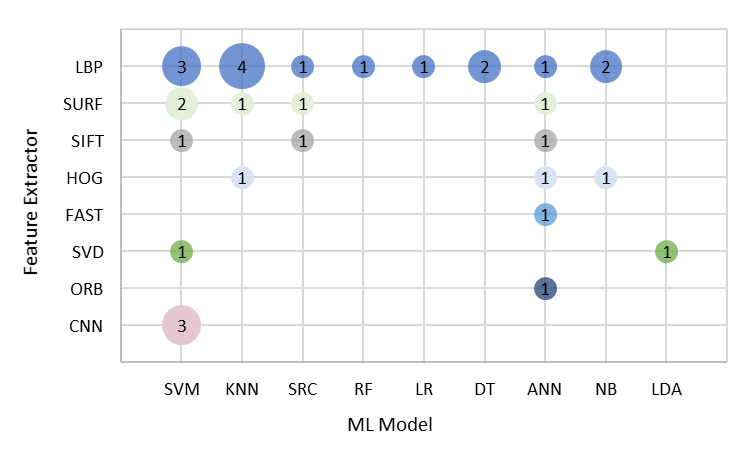}
    \caption{Combination of feature extractors and ML models used for cattle identification in the reviewed studies. The number indicates the number of studies reporting using the combination of Feature Extractor and ML model.}
    \label{fig:FeatureML}
\end{figure}

Fig. \ref{fig:FeatureML} presents a matrix plot considering the combination of feature extraction methods and ML models used for cattle identification. The results show that SVM is mainly used with the CNN, LBP, and SURF feature extraction methods for cattle identification, whereas the KNN and DT classifiers are mainly used with the LBP feature extraction methods. It is also observed that the LBP feature extraction method is used with almost all the ML models for cattle identification. This is mainly because of its powerful texture feature extraction capability from images in pattern recognition systems.

\begin{figure}[t]
    \centering
    \includegraphics[width=.95\textwidth]{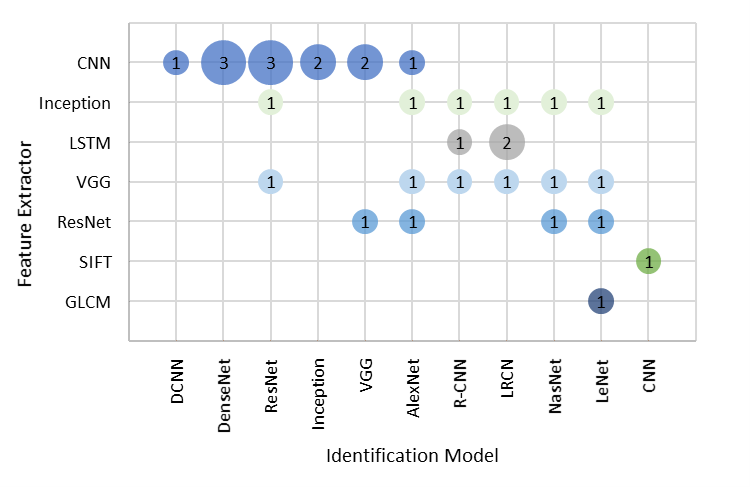}
    \caption{Combination of feature extractors and DL models used for cattle identification. The number indicates the number of studies reporting using the combination of Feature Extractor and DL model}
    \label{fig:FeatureDL}
\end{figure}

CNN and Inception have been widely used in cattle precision farming. They are used as both classifiers and feature extractors. The DL architecture without its last layer (i.e., classifier) is called a feature extractor. The CNN and Inception models excluding all dense layers are performed as feature extractors. Fig. \ref{fig:FeatureDL} shows the combination of feature extraction methods and DL models used for cattle identification in the reviewed studies. It indicates that DenseNet, ResNet and Inception as DL classifiers have mainly been used with CNN as feature extraction methods for cattle identification.   

\subsection{Evaluation metrics}
\begin{figure}[t]
    \centering
    \includegraphics[width=1\textwidth]{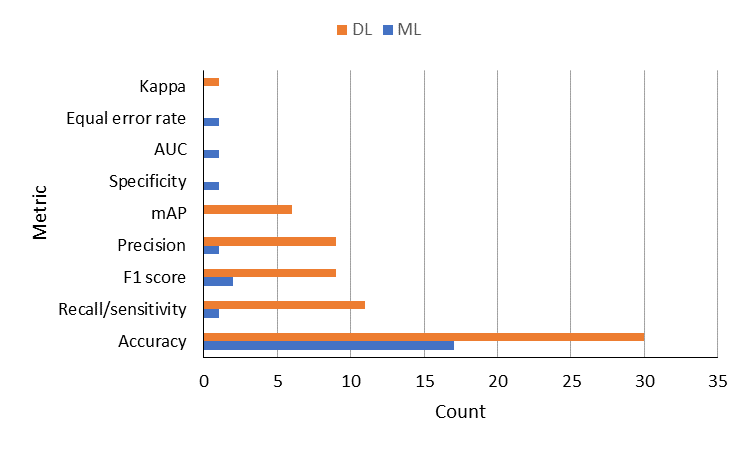}
    \caption{Distribution of performance metrics used in the selected papers.}
    \label{fig:Metrics}
\end{figure}

Different metrics have been used for evaluating the performance of models. To address the fifth research question (RQ5), evaluation metrics are investigated and identified. Nine evaluation metrics were used for cattle identification in the reviewed studies -- accuracy, recall, F1 score, precision, mAP, specificity, the area under the ROC curve (AUC), equal error rate, and Kappa, as shown in Fig. \ref{fig:Metrics}. Accuracy is defined as the percentage of correctly predicted instances and was the most used evaluation metric in ML and DL based studies (17 times for ML and 31 times for DL) for cattle identification. The next most used evaluation metrics were recall, F1 score, and precision. Although 85\% of the reviewed papers used accuracy as the evaluation metric, the recall, F1 score, precision, mAP, and specificity would be best to understand the identification outcomes because these metrics are considered true positive, false positive, true negative, and false negative for evaluating model performance. It is observed that accuracy is the best evaluation metric for individual cattle identification using ML and DL models. It is also noteworthy that mAP and precision are mainly used in DL-based articles as they are appropriate for measuring the DL model accuracy for identification and detection \citep{RRzou2019object}.    

\subsection{Performance of the ML and DL models}
\begin{figure}[ht]
    \centering
    \includegraphics[scale=.78]{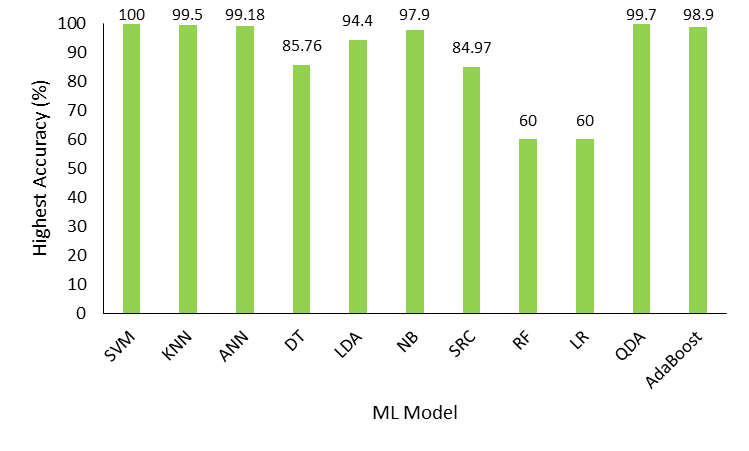}
    \caption{Highest accuracy of ML models used in cattle identification.}
    \label{fig:AccuracyML}
\end{figure}

\begin{figure}[!ht]
    \centering
    \includegraphics[scale=.78]{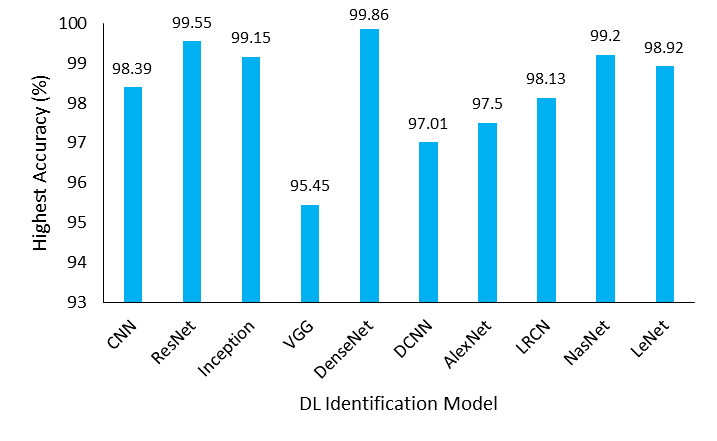}
    \caption{Highest accuracy of DL models used in cattle identification.}
    \label{fig:AccuracyDLindetification}
\end{figure}

\begin{figure}[!ht]
    \centering
    \includegraphics[scale=.7]{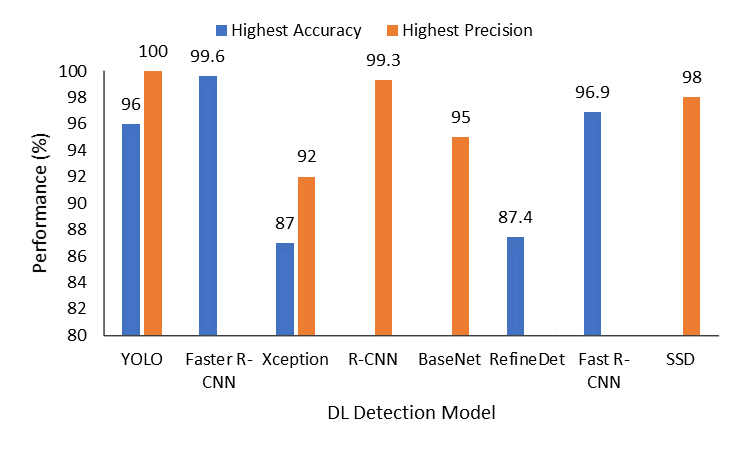}
    \caption{Highest accuracy and precision of DL models used in cattle detection.}
    \label{fig:AccuracyDLdetcection}
\end{figure}

To address the sixth research question (RQ6), the best performance of each model used in the reviewed papers was identified and summarised. Some models are used in many studies. This SLR reported the best one in terms of evaluation metrics. Fig. \ref{fig:AccuracyML} shows the highest accuracy of each ML model used for cattle identification. 
The study found that the top four ML models -- SVM, KNN, ANN, and quadratic discriminant analysis (QDA) had more than 99\% cattle identification accuracy. However, other models also showed good accuracy, such as Naive Bayes (NB), Adaboost, and linear discriminant analysis (LDA). The accuracy of the DL models used for the individual cattle identification is shown in Fig. \ref{fig:AccuracyDLindetification}. The identification models having more than 99\% accuracy are ResNet, Inception, DenseNet, and the neural architecture search network (NasNet). CNN, long-term recurrent convolutional network (LRCN), and LeNet also have good accuracy for cattle identification. As the cattle identification problem included cattle detection in this SLR, the highest accuracy and precision of each DL detection model are identified and shown in Fig. \ref{fig:AccuracyDLdetcection}. Most of the cattle detection papers reported the performance with accuracy and precision for their evaluation metrics. In terms of accuracy and precision, the best detection models are YOLO, Faster-RCNN, and R-CNN. 

\section{Discussions}

\subsection{Challenges and future research directions}
The reviewed articles mentioned different types of challenges encountered during the research. In this SLR, although we identified the challenges for cattle identification using ML and DL approaches, the same challenges can be found for identifying other animals (e.g., sheep and pigs). To address the seventh research question (RQ7), the identified main challenges are discussed in this section, along with future research opportunities.

\textbf{Data quality.} Data quality matters for developing automatic cattle identification systems using ML and DL approaches. In this SLR, poor image quality is one of the main factors for reducing model performance for cattle identification. It was found that illumination variance and motion blur are responsible for noisy images. Additionally, the datasets can be noisy or low quality because they are collected from harsh outdoor and indoor farm environments. As it requires a long time to process the high-quality image datasets, the researchers reduced the image size or divided the images into small pieces to increase the data processing speed \citep{AID47yao2019cow,AID48yang2019dairy}. Several studies in this SLR report that small and unbalanced datasets are the causes of low accuracy \citep{AID10kumar2017muzzle,AID32kumar2018deep,AID34qiao2019individual,AID42guan2020cattle}. Additionally, a few studies stated that nonstandard muzzle print images resulted in low accuracy for ML models. As a larger and complex image or video dataset is significant to train DL models, several studies used the data augmentation method to enhance the training data and their corresponding labels \citep{AID19chen2021angus,AID24andrew2017visual,AID41shen2020individual}. Several research studies identified the redundant information in the video dataset as a challenge for cattle identification \citep{AID30tassinari2021computer,AID34qiao2019individual,AID35de2020recognition}. In these cases, a standard and high-quality cattle dataset is needed to be used for large-scale evaluations of feature extraction methods and cattle detection and identification models.

\textbf{Benchmark dataset.} There is a lack of benchmark datasets that can be used to extract all the relevant features for cattle identification problems. Based on this SLR, we found that a few datasets are publicly available, although the value of these datasets is minimal due to the lack of uniform data standards. As the performance of the existing ML and DL models still depends on the dataset, the standard benchmark dataset needs to be available in the public domain. Thus, the model using a uniform dataset may apply to real firms for cattle identification. A recommendation from the current study is to create a large-scale benchmark cattle dataset that can be used by researchers for cattle identification problems in the field of precision livestock farming.  

\textbf{Data collection duration.} A few studies tested the same identification model with different datasets collected from the same cattle but at different times (days or months). They reported that the model's accuracy was much low when they collected cattle images for training and testing purposes on two different days \citep{AID19chen2021angus}. There is an opportunity for researchers to develop ML or DL models that can handle uncorrelated data from different days and environments.

\textbf{Image overlapping.} Another challenge identified for cattle identification using DL models is that the model results are not satisfactory when the full body or any part of the body overlaps with another animal. Researchers may enhance the DL model's performance by improving the segmentation ability for overlapping individual cattle portions.

\textbf{Feature selection.} Most of the articles reviewed in this SLR used datasets with a small number of individual cattle, therefore there was a lack of efficient feature selection methods on large datasets. Scholars may develop methods to extract important features from the large dataset for cattle identification. Additionally, like human face landmarks, researchers can consider cattle face landmarks as features for cattle face recognition. In this paper, we give a comprehensive overview of cattle identification. For cattle that have coat patterns, for example Holstein, Belgian blue, Shorthorn, Ayrshire and Irish Moiled, the body or whole head area can be used to differentiate animals. For those that have pure colour skin, like Augua, Dexter, and Sussex, the muzzle pattern is the unique biomedical feature to use. Many advanced deep learning models have been developed for object detection. However, only limited models have been applied for cattle identification. There is a large opportunity to continue to develop and promote advanced machine learning technologies for livestock production.

\textbf{Model complexity.} Increasing the number of adjustable weights or parameters in the model architecture increases the complexity of ML and DL models \citep{DISalzubaidi2021review} but may also increase the model accuracy. However, when the model becomes too complex, it tends to overfit the training dataset, and such situations reduce the model performance on the test dataset \citep{DISsrivastava2014dropout,DISxu2019overfitting}. Overfitting was a commonly reported problem in this SLR. It makes the model dataset dependent, and therefore the model does not perform well when applied to other datasets. The researchers use different methods to prevent overfitting problems, including data augmentation, ensembling, and cross-validation. Thus, they can train ML and DL models to generalise well to a new dataset.   

\textbf{Computational cost.} DL models are more computationally expensive than classical ML models, as they require a vast amount of data during the training phase. The amount of computational power required for DL models depends on the size of the dataset as well as the complexity of the model. The researchers stated that over-parameterisation is one of the leading causes of the complex model \citep{DISalzubaidi2021review}. In this SLR, several reviewed studies needed a long time to train the model when they used a large amount of data \citep{AID36andrew2021visual,AID47yao2019cow,AID21phyo2018hybrid}. To reduce the model computational cost, some studies used a transfer learning approach where pre-trained DL models were used for cattle identification. Future researchers can use transfer learning for cattle identification to save training time and improve model performance.

\textbf{Open environment operation challenge.}
Another challenge identified here was the implementation of the cattle identification models in the real farm environment. Although some researchers tested their models on cattle farms, no models were applied to the livestock farm management systems. A large dataset incorporating many individual cattle would be useful to train ML and DL models for application on real cattle farms for identification. Additionally, due to the computation cost of a vision-based system, it is unlikely to run real-time identification without using a server over the cloud. Dairy cattle are often kept indoors, making it relatively easy to implement a vision-based identification system for better biosecurity management. However, for the beef industry, cattle are scattered in an open environment, and reliable and cheap network connectivity is still under development. It will be very costly if images are transferred back to a server via satellite. So the implementation of cattle identification in these environments wild still needs the maturity of the IoT technology. 

At this stage, livestock vision-based tracking and monitoring systems are still under development. There are some proposed static image-based approaches \citep{AID36andrew2021visual,AID22qiao2020bilstm} in use. For example, dairy cattle that are living indoors, where it is easy to set up the vision system. But for cattle scattered in the field, it will be very challenging to obtain consistent features for accurate object recognition from the captured images/videos. Therefore, the power of machine learning and computer vision has not been fully integrated into practice.

\subsection{Limitations of this study}
The main limitation of this study is the database search for the relevant articles. This SLR considered four electronic databases. We identified 55 studies using the search strategy described in the methodology section. There may be more articles available that have not been included in this study due to not considering other electronic databases. Some relevant articles might have been missed because of the search keyword string. To maximise the number of relevant articles, we chose the databases by considering the scope of this study, and we conducted a broad search with a search string including main search terms and their synonyms. Another possible limitation is the data extraction from the selected articles, because some data could be missed in this process. To minimise the amount of missing data, we cross-checked the analysed data. These limitations will provide doors for future SLR, but they do not inhibit the main goal of presenting a comprehensive picture of the usage of ML and DL approaches for cattle identification.

\section{Conclusions}
This study presents a systematic literature review of the application of classical ML and DL models for cattle identification and detection in cattle farming. Significant numbers of papers were investigated, and the research value of automatic cattle identification using ML and DL approaches has been extensively examined. In this SLR, we have determined several important insights such as datasets, visual cattle features, feature extraction methods, ML and DL algorithms, performance evaluation metrics, and challenges related to the use of ML and DL in this field. The results show that the application of ML and DL approaches for automatic cattle identification has gained popularity among researchers in recent years because of their ability to learn unique features and provide high accuracy. Based on this SLR, no conclusion can be made about the best model. However, it is observed that some ML and DL models are frequently used for cattle identification. The most used ML models are SVM, KNN, and ANN, whereas the most used DL models are CNN, ResNet, Inception, YOLO, and Faster R-CNN. Although other ML and DL models are also used for cattle identification. This SLR does not compare the performance of different types of ML and DL models used in cattle identification, as the results are case sensitive due to the sparsity of the benchmark dataset. Thus, there is no commonly accepted evaluation scheme for a fair comparison. The main challenges identified in this study for the application of ML and DL to cattle identification include dataset quality and availability, benchmark datasets, model selection and complexity, as well as real-time cattle identification in the farm environment. The current major challenges need to be addressed for efficient and effective cattle identification systems and further improvements to ML and DL models. It is concluded that automated and real-time cattle identification systems will play an important role in livestock farm management in the future. We believe that this study will facilitate future researchers in developing automatic cattle identification systems.  

\section*{Declarations of interest}
The authors declare that they have no known competing financial interests or personal relationships that could have appeared to influence the work reported in this paper.

\section*{Acknowledgements}
We thank Will Swain (TerraCipher) for his valuable comments.

\section*{Funding}
This project was supported by funding from Food Agility CRC Ltd, funded under the Commonwealth Government CRC Program. The CRC Program supports industry-led collaborations between industry, researchers and the community.







  \bibliographystyle{elsarticle-harv}
  \bibliography{Reference}





\end{document}